\documentclass{article}

\usepackage{amsmath}
\usepackage{amssymb}
\usepackage{amsopn}
\usepackage{algorithm}
\usepackage{algorithmic}
\usepackage{epsfig}
\usepackage{microtype}
\usepackage{graphicx}
\usepackage{subfigure}
\usepackage{booktabs}
\usepackage{hyperref}

\newenvironment{proof}{\paragraph{Proof:}}{\hfill$\square$}

\newtheorem{th_innerprod_kern}{Theorem}
\newtheorem{mercer_theorem}[th_innerprod_kern]{Theorem}
\newtheorem{mercer_extend}[th_innerprod_kern]{Theorem}

\newtheorem{th_kernel_weight}[th_innerprod_kern]{Theorem}

\newtheorem{def_2qkernel}{Definition}
\newtheorem{def_freekernel}[def_2qkernel]{Definition}
\newtheorem{def_normfreekernel}[def_2qkernel]{Definition}

\newcommand{\perm}[0]{{\rm perm}}
\newcommand{\tsp}[0]{{\rm T}}

\DeclareMathOperator\sgn{sgn}

\DeclareMathOperator\gp{GP} 
\DeclareMathOperator\argmax{argmax} 
\DeclareMathOperator\predataset{\diamond} 

\begin{document}

\title{Covariance Function Pre-Training with $m$-Kernels for Accelerated Bayesian Optimisation}
\author{Alistair Shilton\footnote{Prada, Deakin University, Waurn Ponds, Australia.  Correspondence to: Alistair Shilton $<$alistair.shilton@deakin.edu.au$>$}, 
Sunil Gupta${}^*$, 
Santu Rana${}^*$, 
Pratibha Vellanki${}^*$, \\
Cheng Li${}^*$, 
Laurence Park\footnote{Western Sydney University}, 
Svetha Venkatesh${}^*$, 
Alessandra Sutti\footnote{Institut for Frontier Materials, Deakin University, Waurn Ponds, Australia}, \\
David Rubin${}^{\ddagger}$, 
Thomas Dorin${}^{\ddagger}$, 
Alireza Vahid${}^{\ddagger}$, \\
Murray Height${}^{\ddagger}$, 
Teo Slezak${}^{\ddagger}$}

\maketitle

\begin{abstract}

The paper presents a novel approach to direct covariance function learning for 
Bayesian optimisation, with particular emphasis on experimental design 
problems where an existing corpus of condensed knowledge is present.  The 
method presented borrows techniques from reproducing kernel Banach space 
theory (specifically $m$-kernels) and leverages them to convert (or re-weight) 
existing covariance functions into new, problem-specific covariance 
functions.  The key advantage of this approach is that rather than relying on 
the user to manually select (with some hyperparameter tuning and 
experimentation) an appropriate covariance function it constructs the 
covariance function to specifically match the problem at hand.  The technique 
is demonstrated on two real-world problems - specifically alloy design and 
short-polymer fibre manufacturing - as well as a selected test function.

This research was supported by the Australian Research Council World Class Future
Fibre Industry Transformation Research Hub (IH140100018)

\end{abstract}

\section{Introduction}

Patents and technical handbooks contain condensed knowledge. That is, they 
contain the distilled version of knowledge that bubbles up from a much larger 
body of supporting work often available as technical publications. This 
condensed knowledge is extremely relevant to a new experimental design process 
but the knowledge in most part cannot be directly applied because 
experimenters have to extract insights and meta-knowledge and then transfer it 
to the new experimental set up. For example, in metallurgy patents, we can 
extract the state of the art in terms of alloy compositions and heat 
treatments, and how it relates to properties such as yield strength. More 
generally in many situations we may have a good amount of data clustered 
around a set of ``well-known'' regions in experimental space and a dearth of 
data elsewhere.

In this paper we present a technique that makes use of any condensed knowledge 
we may have that is relevant to the problem to accelerate the experimental 
design process using a Bayesian optmisation framework. One difficulty with 
Bayesian optimisation using Gaussian process models in practice is that of 
covariance function selection.  In principle one assumes that data is drawn 
from a distribution whose covariance function matches that of the covariance 
function used by the Gaussian process model.  Typically this will take the 
form of a squared-exponential function, Mat{\'e}rn kernel or similar, where 
the relevant hyperparameter(s) are selected to maximise the log-likelihood (or 
similar) fitness criteria.  However it is by no means guaranteed that this 
covariance function represents a good approximation of 
the true covariance function.  In light of this, the question that naturally 
arises is: using condensed knowledge, how can we (automatically) find a 
covariance function that better fits the actual covariance?  This more closely 
fitted covariance function allows us to build more accurate Gaussian process 
models, which we use to speed up the Bayesian optimisation process.

We approach this problem from the perspective of kernel methods (a kernel 
being a covariance function by another name).  To do this we begin with the 
theory of $m$-kernels, where an $m$-kernel is a map $K : \mathbb{X}^m \to 
\mathbb{R}$ that arises as a special case of reproducing kernel Banach space 
theory \cite{Der1,Fas1,Zha11}.  In particular, an $m$-kernel is an 
$m$-semi-inner-product \cite{Dag1} evaluated on the $m$ arguments mapped into 
some feature space, which in the case $m=2$ reduces to a standard (Mercer) 
kernel (covariance function).  We show that, for a particular subset of 
$m$-kernels we call free $m$-kernels, the $m$-kernels have the useful property 
that the feature map they implicitly embody is independent of $m$.  Using this 
observation we show how the feature map itself may be directly modified to 
give weight to relevant features and less to irrelevant ones; and moreover 
that this weighting may be learned directly from condensed datasets such as 
patents  and handbooks.

We demonstrate our results using two real world settings: \emph{first} we 
design a new hybrid Aluminium alloy based on 46 existing patents for 
aluminium 6000, 7000 and 2000 series. We learn the kernel from this condensed 
set of patents  and use it to accelerate the design of the new aluminium 
alloy. \emph{Second} we test our algorithm for a new short polymer fiber 
design using micro-fluid devices. Experimental data is available from 
\emph{Device A} - a particular geometric configuartion using a gear pump. We 
learn the kernel using this data and then use it to design short polymer fiber 
on a \emph{Device B} -  a different configuration using a lobe pump. We 
demonstrate in both our experiments that our method outperforms standard 
Bayesian optimisation for several acquisition functions.
    
Our main contributions are:
\begin{itemize}
 \item Extension of Mercer's Theorem (Mercer's condition) to $2q$-kernels 
       (theorem \ref{th:mercer_extend}, section \ref{sec:mkstruct}).
 \item Definition of free kernels: families of $m$-kernels whose corresponding 
       (implied) feature weights and maps are independent of $m$ (definition 
       \ref{def:def_freekernel}, section \ref{sec:freekern}).
 \item Definition of $m$-semi-inner-product kernels ($m$-kernel analogues of 
       dot product kernels) and proof that these are free kernels
       (theorem \ref{th:th_innerprod_kern}, section \ref{sec:makefreekern}).
 \item Definition of $m$-kernel analogues of linear, polynomial, exponential 
       and squared-exponential kernels (table \ref{table:kern_table}).
 \item Introduction of kernel re-weighting, a method of tuning free kernels to 
       adjust implied feature weights (theorem \ref{th:th_kernel_weight}, 
       section \ref{sec:kernreweight}).
 \item A modified Bayesian optimisation algorithm that leverages kernel 
       re-weighting to pre-tune the covariance function using additional data 
       to better match the problem at hand (algorithm \ref{alg:modded_bbo}, 
       section \ref{sec:modbo}).
 \item Demonstration of the utility for accelerated optimisation in two practical problems, 
       namely short polymer fibre production and aluminium alloy design
       (section \ref{sec:results}).
\end{itemize}

\subsection{Notation}

Sets are written $\mathbb{A}, \mathbb{B}, \ldots$; and $\mathbb{N}^+ = \{ 1, 
2, \ldots \}$, $\mathbb{N}_n = \{ 1, 2, \ldots, n \}$.  If $\mathbb{I} = \{ 
i_1, i_2, \ldots, i_n \}$ then $\perm ( \mathbb{I} )$ is the set of all 
permutations ($n$-tuples) of these indices.  Vectors are bold lower case ${\bf 
a}, {\mbox{\boldmath $\tau$}}$.  Element $i$ of vector ${\bf a}$ is $a_i$.  
${\bf a} {\bf b}$ is the elementwise product, ${\bf a}^{\rm T} {\bf b} = 
\left< {\bf a}, {\bf b} \right>$ the inner product, ${\bf a}^b$ the 
elementwise power, ${\bf 1}$ a vector of $1$s, and ${\bf 0}$ a vector of 
$0$s.  The $m$-semi-inner-product $\ll\! \ldots \!\gg_m : (\mathbb{R}^n)^m \to 
\mathbb{R}$ is the multi-linear map \cite{Dag1}:
\[
 \begin{array}{rl}
  \ll\! {\bf a}, {\bf a}', \ldots, {\bf a}^{\ldots} \!\gg_m 
  &\!\!\!= \left< {\bf 1}, {\bf a} {\bf a}' \ldots {\bf a}^{\ldots} \right> \\
 \end{array}
\]

\section{Problem Statement and Background}

We want to maximise a function $f : \mathbb{R}^n \to \mathbb{R}$ that is 
expensive to evaluate to find ${\bf x}^* = \argmax_{{\bf x} \in \mathbb{R}^n} 
f ({\bf x})$.  It is assumed that $f \sim \gp ( 0, K )$ is a draw from a zero 
mean {G}aussian process \cite{Ras1} with covariance function $K : \mathbb{R}^n 
\times \mathbb{R}^n \to \mathbb{R}$.  We note that in most realistic use-cases 
$K$ is not given a-priori but rather obtained using a combination of 
experience, assumptions and heuristics such as max-log-likelihood to select a 
``good-enough'' covariance function from a set of well-known covariance 
functions (squared exponential, Mat{\'e}rn, etc).

To accelerate the optimisation we are given an additional dataset 
$\mathcal{D}_{\predataset} = \{ ( {\bf x}_{{\predataset} i}, y_{{\predataset} 
i} ) | {\bf x}_{{\predataset} i} \in \mathbb{R}^n, y_{{\predataset} i} = 
f_{\predataset} ( {\bf x}_{{\predataset} i} ) + \epsilon_{{\predataset} i} \in 
\mathbb{Y} \}$, where $\epsilon_{{\predataset} i}$ is a noise term.  This 
data may be derived from for example patents, handbooks, datasheets of similar 
sources of (relevant) domain knowledge.  In general $f_{\predataset} \ne f$ 
and $\mathbb{Y} \ne \mathbb{R}$, but we assume that $f_{\predataset}$ and $f$ 
are similar in the following sense: if we were to use some machine learning 
algorithm to fit the models:
\[
 \begin{array}{rl}
  g_t             \left( {\bf x} \right) &\!\!\!= \left< {\bf w}_t,             {\mbox{\boldmath $\varphi$}} \left( {\bf x} \right) \right> + b_t       \\
  g_{\predataset} \left( {\bf x} \right) &\!\!\!= \left< {\bf w}_{\predataset}, {\mbox{\boldmath $\varphi$}} \left( {\bf x} \right) \right> + b_{\predataset} \\
 \end{array}
\]
to $\mathcal{D}_{\predataset}$ and $\mathcal{D}_t$, respectively, where 
$\mathcal{D}_t = \{ ( {\bf x}_i, y_i ) | {\bf x}_i \in \mathbb{R}^n, y_i = f ( 
{\bf x}_i ) + \epsilon_i \in \mathbb{R}, \epsilon_i \sim \mathcal{N} ( 0, 
\nu^2 ) \}$ represents data generated by $f$, then the relative magnitudes of 
the feature weight vectors ${\bf w}_t$ and ${\bf w}_{\predataset}$ would be 
similar - that is (loosely speaking), if $|w_{t i}|/\|{\bf w}_t\|_2$ is large 
(small) then $|w_{{\predataset} i}|/\|{\bf w}_{\predataset}\|_2$ is also large 
(small).  We will use $\mathcal{D}_{\predataset}$, and in particular the 
weights ${\bf w}_{\predataset}$ (or rather their representation) obtained 
using a support vector machine, to directly construct a covariance function 
$K$ fitted to the objective $f$.

\subsection{Bayesian Optimisation}

Bayesian optimisation \cite{Bro2} (BO) is a technique for maximising expensive 
(to evaluate) black-box functions $f$ in the fewest iterations possible.  
Bayesian optimisation works by maintaining a GP model of $f$ constructed from 
the set $\mathcal{D}_{t-1}$ of observations of $f$ up to (but not including) the 
current iteration $t$.  Based on this model, the acquisition function $a_t$, 
which is cheap to evaluate and depends on our model of $f$, is maximised to 
find the next point ${\bf x}_t$ to be evaluated.  This point is evaluated to 
obtain $y_t = f ({\bf x}_t) + \epsilon$, the model is updated, and the 
algorithm continues.  A typical Bayesian optimisation algorithm is given in 
algorithm \ref{alg:generic_bbo}.

\begin{algorithm}
\caption{Generic Bayesian Optimisation}
\label{alg:generic_bbo}
\begin{algorithmic}
 \INPUT $\mathcal{D}_0 = \{ ( {\bf x}, y ) | y = f ( {\bf x} ) + \epsilon, \epsilon \sim \mathcal{N} ( 0, \nu^2 ) \}, K$.
 \STATE Proceed modelling $f \sim \gp ( 0, K )$.
 \FOR{$t=1,2,\ldots,T$}
 \STATE Select test point ${\bf x}_t = \argmax_x a_t ({\bf x})$.
 \STATE Perform Experiment $y_t = f ( {\bf x}_t ) + \epsilon_t, \epsilon_t \sim \mathcal{N} ( 0, \nu^2 )$.
 \STATE Update $\mathcal{D}_t := \mathcal{D}_{t-1} \cup \{ ({\bf x}_t,y_t) \}$.
 \ENDFOR
\end{algorithmic}
\end{algorithm}

As $f \sim \gp ( 0, K )$ is a draw from a zero mean {G}aussian process 
\cite{Ras1}, given observations $\mathcal{D}_t = \{ ( {\bf x}_i, y_i ) | {\bf 
x}_i \in \mathbb{R}^n, y_i = f ( {\bf x}_i ) + \epsilon_i \in \mathbb{Y}, 
\epsilon_i \sim \mathcal{N} ( 0, \nu^2 ) \}$, the posterior $f ( {\bf x} ) | 
\mathcal{D}_t \sim \mathcal{N} ( \mu_t ({\bf x}), \sigma_t^2 ({\bf x}) )$ has 
mean and variance:
\begin{equation}
 \begin{array}{rl}
  \mu_t      \left( {\bf x} \right) &\!\!\!= {\bf k}_t^{\rm T} \left( {\bf x} \right) \left( {\bf K}_t + \nu^2 {\bf I} \right)^{-1} {\bf y}_t \\
  \sigma_t^2 \left( {\bf x} \right) &\!\!\!= k \left( {\bf x},{\bf x} \right) - {\bf k}_t^{\rm T} \left( {\bf x} \right) \left( {\bf K}_t + \nu^2 {\bf I} \right)^{-1} {\bf k}_t \left( {\bf x} \right) \!\!\!\\
 \end{array}
 \label{eq:defmusigma}
\end{equation}
where ${\bf y}_t = [ y_i ]_{(-,y_i) \in \mathcal{D}_t}$, ${\bf k}_t ( {\bf x} 
) = [ K ({\bf x}_i,{\bf x}) ]_{({\bf x}_i,-) \in \mathcal{D}_t}$ and ${\bf 
K}_t = [ K ({\bf x}_i,{\bf x}_j) ]_{({\bf x}_i,-)}$ ${}_{,({\bf x}_j,-) \in 
\mathcal{D}_t}$.\footnote{We say $({\bf x},-) \in \mathcal{D}_t$ if $\exists y 
\in \mathbb{R} : ( {\bf x}, y ) \in \mathcal{D}_t$ and $( -,y) \in 
\mathcal{D}_t$ if $\exists {\bf x} \in \mathbb{R}^n : ( {\bf x}, y ) \in 
\mathcal{D}_t$.}  Using this model typical acquisition functions include 
probability of improvement (PI) \cite{Kus1}, expected improvement (EI) 
\cite{Jon1}, GP upper confidence bound (GP-UCB) \cite{Sri1} and predictive 
entropy search (PES) \cite{Her3}:
\[
 \begin{array}{rl}
  a_t^{\rm  PI} \left( {\bf x} \right) &\!\!\!= \Phi \left( Z \right) \\
  a_t^{\rm  EI} \left( {\bf x} \right) &\!\!\!= \left( \mu_{t-1} \left( {\bf x} \right) - y^+ \right) \Phi \left( Z \right) + \sigma_{t-1} \left( {\bf x} \right) \phi \left( Z \right) \\
  a_t^{\rm UCB} \left( {\bf x} \right) &\!\!\!= \mu_{t-1} \left( {\bf x} \right) + \sqrt{\beta_t} \sigma_{t-1} \left( {\bf x} \right) \\
 \end{array}
\]
where $Z = \frac{\mu_{t-1} ({\bf x}) - y^+}{\sigma_{t-1} ({\bf x})}$, $y^+ = 
\max y_i$ and $\beta_t$ is a sequence of constants \cite{Sri1}.

\subsection{$m$-Kernels and Support Vector Machines}

Another name for a covariance function that is common in the machine learning 
literature is kernel function.  A (Mercer) kernel function is any function $K 
: \mathbb{R}^n \times \mathbb{R}^n \to \mathbb{R}$ for which there exists a 
corresponding (implicitly defined) feature map ${\mbox{\boldmath $\varphi$}} : 
\mathbb{R}^n \to \mathbb{R}^d$ such that $\forall {\bf x},{\bf x}' \in 
\mathbb{R}^n$:
\[
 \begin{array}{l}
  K \left( {\bf x},{\bf x}' \right) = \left< {\mbox{\boldmath $\varphi$}} \left( {\bf x} \right), {\mbox{\boldmath $\varphi$}} \left( {\bf x}' \right) \right>
 \end{array}
\]
Equivalently, if we define $\theta ({\bf x}, {\bf x}')$ to be the angle 
between ${\mbox{\boldmath $\varphi$}} ({\bf x})$ and ${\mbox{\boldmath 
$\varphi$}} ({\bf x}')$ in feature space:
\begin{equation}
 \begin{array}{rl}
  K \left( {\bf x},{\bf x}' \right) 
  &\!\!\!= \left\| {\mbox{\boldmath $\varphi$}} \left( {\bf x} \right) \right\|_2 \left\| {\mbox{\boldmath $\varphi$}} \left( {\bf x}' \right) \right\|_2 \cos \theta \left( {\bf x}, {\bf x}' \right) \\
  &\!\!\!= \sqrt{ K \left( {\bf x}, {\bf x} \right) } \sqrt{ K \left( {\bf x}', {\bf x}' \right) } \cos \theta  \left( {\bf x}, {\bf x}' \right) \\
 \end{array}
 \label{eq:anglesandstuff}
\end{equation}
Thus the kernel (covariance function) $K$ provides a measure of the similarity 
of ${\bf x}$ and ${\bf x}'$ in terms of their alignment in feature space.  For 
kernel such as the SE kernel for which $K({\bf x}, {\bf x}) = 
1$ this simplifies to $K ( {\bf x}, {\bf x}' ) = \cos \theta ({\bf x}, {\bf 
x}')$.

An extension that arises in the context of generalised norm SVMs 
\cite{Man50,Man51} and reproducing kernel Banach-space (RKBS) theory 
\cite{Der1,Fas1,Zha11} is the $m$-kernel.\footnote{Also called a moment 
function in \cite{Der1}.}  Formally:
\begin{def_2qkernel}
 For $m \in \mathbb{N}^+$ an {\em $m$-kernel} is a function $K : 
 (\mathbb{R}^n)^m \to \mathbb{R}$ for which there exists a feature map 
 ${\mbox{\boldmath $\varphi$}} : \mathbb{R}^n \to \mathbb{R}^d$ such that, 
 $\forall {\bf x}, {\bf x}', \ldots, {\bf x}^{\ldots} \in \mathbb{R}^n$:
 \begin{equation}
  \begin{array}{l}
   \!\!\!\!\!\!K \!\!\left( {\bf x}\!, {\bf x}'\!, \ldots, {\bf x}^{\ldots} \right) 
   \!=\! \;\ll\! {\mbox{\boldmath $\varphi$}} \!\left( {\bf x} \right)\!, {\mbox{\boldmath $\varphi$}} \!\left( {\bf x}' \right)\!, \ldots, {\mbox{\boldmath $\varphi$}} \!\left( {\bf x}^{\ldots} \right) \!\gg_m\!\!\!\!\!\!
  \end{array}
  \label{eq:2qkern}
 \end{equation}
 where $\ll\! \ldots \!\gg_m$ is an $m$-semi-inner-product \cite{Dag1} of the 
 form:
 \[
  \begin{array}{l}
   \ll\! {\mbox{\boldmath $\phi$}},  {\mbox{\boldmath $\phi$}}', \ldots, {\mbox{\boldmath $\phi$}}^{\ldots} \!\gg_m 
   = \left< {\bf 1}, {\mbox{\boldmath $\phi$}} {\mbox{\boldmath $\phi$}}' \ldots {\mbox{\boldmath $\phi$}}^{\ldots} \right>
  \end{array}
 \]
 and ${\mbox{\boldmath $\phi$}} {\mbox{\boldmath $\phi$}}' \ldots 
 {\mbox{\boldmath $\phi$}}^{\ldots}$ is the elementwise product.  We say that 
 feature map ${\mbox{\boldmath $\varphi$}}$ satisfying (\ref{eq:2qkern}) is 
 {\em implied} by $K$.
\label{def:def_2qkernel}
\end{def_2qkernel} 

Note that a $2$-kernel is just a standard (Mercer) kernel.  A canonical 
application of $m$-kernels is the $p$-norm support vector machine (SVM) (aka 
the max-margin $L^p$ moment classifier \cite{Der1}).  Let 
$\mathcal{D} = \{ ( {\bf x}_{i}, y_{i}) | {\bf x}_{i} \in \mathbb{R}^n, y_{i} 
\in \mathbb{R} \}$ be a regression training set.\footnote{A similar procedure 
applies for binary classification etc. - see supplementary material.}  We wish to (implicitly) find a sparse 
(in ${\bf w}$) function:\footnote{The weight ${\bf w}$ is (almost) never 
calculated explicitly but rather defined implicitly in the dual form.}
\[
 \begin{array}{l}
  g \left( {\bf x} \right) = \left< {\bf w}, {\mbox{\boldmath $\varphi$}} \left( {\bf x} \right) \right> + b
 \end{array}
\]
to fit the data.  The $p$-norm SVM is one approach to this problem.  Letting 
$1 < p \leq 2$ be dual to $2q$, $q \in \mathbb{N}^+$ (i.e. $1/p + 1/2q = 1$) 
the $p$-norm SVM primal is:
\begin{equation}
 \begin{array}{l}
  \begin{array}{l}
   \mathop{\min}\limits_{{\bf w},b,\xi} \frac{1}{p} \left\| {\bf w} \right\|_p^p + \frac{C}{N} \sum_i \xi_i
  \end{array} \\
  \begin{array}{ll}
   \mbox{such that:} & \left| {\bf w}^\tsp {\mbox{\boldmath $\varphi$}} \left( {\bf x}_{i} \right) + b - y_{i} \right| \leq \epsilon + \xi_i \; \forall i \\
   & \xi_i \geq 0 \; \forall i \\
  \end{array} \\
 \end{array}
 \label{eq:theprimal_pnorm}
\end{equation}
Using the $m$-kernel trick (see supplementary material) the dual form of the 
$p$-norm SVM may be derived:
\begin{equation}
 \!\!\!\!\!\!\!\!\!\!\!\!\begin{array}{l}
  \begin{array}{rl}
   \mathop{\min}\limits_{\alpha} &\!\!\!\frac{1}{2q} \mathop{\sum}\limits_{i_1,i_2,\ldots,i_{2q}} \!\!\!\alpha_{i_1} \alpha_{i_2} \ldots \alpha_{i_{2q}} K_{i_1, i_2, \ldots, i_{2q}} + \ldots \\
                                 &\!\!\!\ldots + \epsilon \sum_i \left| \alpha_i \right| - \sum_i y_{i} \alpha_i
  \end{array} \\
  \begin{array}{ll}
   \mbox{such that:} & -\frac{C}{N} \leq \alpha_i \leq \frac{C}{N} \; \forall i \\
   & \sum_i \alpha_i = 0 \\
  \end{array} \\
 \end{array}\!\!\!\!\!\!\!\!\!\!\!\!
 \label{eq:thedual}
\end{equation}
where $K_{i_1, i_2, \ldots, i_{2q}} = K ( {\bf x}_{i_1}, {\bf x}_{i_2}, 
\ldots, {\bf x}_{i_{2q}} )$ and $K$ is a $2q$-kernel (in practice we start 
with $K$ and never need to know the feature map ${\mbox{\boldmath $\varphi$}}$ 
implied by $K$).  Moreover:
\begin{equation}
 \begin{array}{l}
  g \left( {\bf x} \right) = \!\!\!\!\!\mathop\sum\limits_{i_2, \ldots, i_{2q}}\!\!\!\!\! \alpha_{i_2} \ldots \alpha_{i_{2q}} K \left( {\bf x}, {\bf x}_{i_2}, \ldots, {\bf x}_{i_{2q}} \right) + b
 \end{array}
 \label{eq:dual_p_trained}
\end{equation}
and by representor theory the (implicitly defined) weight vector ${\bf w}$ is:
\begin{equation}
 \begin{array}{l}
  {\bf w} = \sum_i \alpha_i {\mbox{\boldmath $\varphi$}} \left( {\bf x}_{i} \right)
 \end{array}
 \label{eq:w_rep}
\end{equation}

\section{The Structure of $m$-Kernels} \label{sec:mkstruct}

Before proceeding to our main contribution we must first establish some 
preliminary results.  We begin by presenting theoretical conditions that must 
be met by a function $K : (\mathbb{R}^n)^{m} \to \mathbb{R}$ to be a $m$-kernel 
under definition \ref{def:def_2qkernel}.  As we are primarily interested in 
$2q$-kernels we will largely focus on this case.  The following provides an 
analogue of Mercer's condition \cite{Mer1} for $2q$-kernels:
\begin{mercer_extend}[$2q$-kernel analogue of Mercer's theorem]
 Let $K : (\mathbb{R}^n)^{2q} \to \mathbb{R}$, $q \in \mathbb{N}^+$, be a 
 continuous function that is symmetric under the exchange of any two arguments 
 and such that $L : (\mathbb{R}^n)^q \times (\mathbb{R}^n)^q \to \mathbb{R}$ 
 defined by:
 \[
  \!\!\!\!\!\!\begin{array}{l}
   L \!\left( \!\left[ \!\!\begin{array}{c} {\bf x} \\ \vdots \\ {\bf x}^{\ldots} \\ \end{array}\!\! \right]\!, \left[ \!\!\begin{array}{c} {\bf z} \\ \vdots \\ {\bf z}^{\ldots} \\ \end{array}\!\! \right]\! \right)\! = K \left( {\bf x}, \ldots, {\bf x}^{\ldots}, {\bf z}, \ldots, {\bf z}^{\ldots} \right)
  \end{array}\!\!\!\!\!\!
 \]
 satisfies Mercer's condition.  Then there exists ${\mbox{\boldmath $\varphi$}} 
 : \mathbb{R}^n \to \mathbb{R}^d$, $d \in \mathbb{N}^+ \cup \{ \infty \}$, such 
 that:
 \[
  \begin{array}{l}
   \!\!\!\!\!\!K \left( {\bf x}, {\bf x}', \ldots, {\bf x}^{\ldots} \right) 
   = \;\ll\! {\mbox{\boldmath $\varphi$}} \left( {\bf x} \right), {\mbox{\boldmath $\varphi$}} \left( {\bf x}' \right), \ldots, {\mbox{\boldmath $\varphi$}}\! \left( {\bf x}^{\ldots} \right) \!\gg_{2q}\!\!\!\!\!\!
  \end{array}
 \]
 the series being uniformly and absolutely convergent. 
 \label{th:mercer_extend}
\end{mercer_extend}
\begin{proof}
See supplementary material.
\end{proof}

\subsection{$m$-Kernels and Free Kernels} \label{sec:freekern}

For the purposes of the present paper we define a special classes of 
$m$-kernels, namely the free kernels:
\begin{def_freekernel}
 A {\em free kernel} is a family of functions $K_m : (\mathbb{R}^n)^m \to 
 \mathbb{R}$ indexed by $m \in \mathbb{N}^+$ for which there exists a feature 
 map ${\mbox{\boldmath $\vartheta$}} : \mathbb{R}^n \to \mathbb{R}^d$ and 
 feature weights ${\mbox{\boldmath $\tau$}} \in \mathbb{R}^d$, {\em both 
 independent of $m$,} such that $\forall m \in \mathbb{N}^+$:
 \begin{equation}
  \begin{array}{l}
   \!\!\!\!\!\!K_m \left( {\bf x}, \ldots, {\bf x}^{\ldots} \right) 
   = \;\ll\! {\mbox{\boldmath $\tau$}}^2, {\mbox{\boldmath $\vartheta$}} \left( {\bf x} \right)\!, \ldots, {\mbox{\boldmath $\vartheta$}} \left( {\bf x}^{\ldots} \right) \!\gg_{m+1}\!\!\!\!\!\!
  \end{array}
  \label{eq:freekern}
 \end{equation}
\label{def:def_freekernel}
\end{def_freekernel} 

Note that for fixed $m$ a free kernel defines (is) an $m$-kernel:
\[
 \begin{array}{l}
  \!\!\!\!\!\!K_m \left( {\bf x}, \ldots, {\bf x}^{\ldots} \right) 
  = \;\ll\! {\mbox{\boldmath $\varphi$}}_m \left( {\bf x} \right)\!, \ldots, {\mbox{\boldmath $\varphi$}}_m \left( {\bf x}^{\ldots} \right) \!\gg_{m}\!\!\!\!\!\!
 \end{array}
\]
with implied feature map:
\begin{equation}
 \begin{array}{l}
  {\mbox{\boldmath $\varphi$}}_m \left( {\bf x} \right) = {\mbox{\boldmath $\tau$}}^{2/m} {\mbox{\boldmath $\vartheta$}} \left( {\bf x} \right)
 \end{array}
 \label{eq:mkernfreemap}
\end{equation}
where ${\mbox{\boldmath $\tau$}}^{2/m}$ is an the elementwise power and 
${\mbox{\boldmath $\tau$}}^{2/m} {\mbox{\boldmath $\vartheta$}} ({\bf x})$ an 
elementwise product.  Note that, while all free kernels define $m$-kernels, 
not all $m$-kernels derive from free kernels for given $m$.  

We also define normalised free kernels:
\begin{def_normfreekernel}
 Let $K_m : (\mathbb{R}^n)^m \to \mathbb{R}$ be a free kernel.  Then for a 
 given $m$ the function $\hat{K}_m : (\mathbb{R}^n)^m \to \mathbb{R}$ defined 
 by:
 \[
  \!\!\!\!\!\!\begin{array}{l}
   \hat{K}_m \!\left( {\bf x}_1, \ldots, {\bf x}_m \!\right) = \left( \!\mathop{\prod}\limits_{i \in \mathbb{N}_m} \!\!\!\frac{1}{\sqrt{\!K_2 \left( {\bf x}_i, {\bf x}_i \right)}} \!\right) \!\!K_m \!\left( {\bf x}_1, \ldots, {\bf x}_m \right)
  \end{array}\!\!\!\!\!\!
 \]
 is a free kernel, which we call the {\em normalised} free kernel.
\label{def:def_normfreekernel}
\end{def_normfreekernel} 
The standard SE ($2$-)kernel is an example of a normalised $2$-kernel as 
it may be constructed by normalising the exponential kernel $K ({\bf x},{\bf 
x}') = \exp (\frac{1}{\sigma}\!\!<\!\!{\bf x},{\bf x}'\!\!>\!)$.  It is 
straightforward to show that if ${\mbox{\boldmath $\tau$}}$ and 
${\mbox{\boldmath $\vartheta$}}$ are the feature weights and feature map 
implied by the free kernel $K_m$ then the normalised free kernel $\hat{K}_m$ 
implies feature weights $\hat{\mbox{\boldmath $\tau$}} = {\mbox{\boldmath 
$\tau$}}$ and feature map $\hat{\mbox{\boldmath $\vartheta$}} ({\bf x}) = (\| 
{\mbox{\boldmath $\tau$}} {\mbox{\boldmath $\vartheta$}} ({\bf x})\|_2)^{-1} 
{\mbox{\boldmath $\vartheta$}} ({\bf x})$

\subsection{Constructing Free Kernels} \label{sec:makefreekern}

We now consider the question of constructing free kernels.  In RKHS theory 
there exist a range of ``standard'' kernels ($2$-kernels) from which to 
choose.  In \cite{Der1} it is shown that any $2$-kernel $K$ may be used to 
construct a $2q$-kernel ${\bar{K}}$ using the symmetrised product:
\[
 \begin{array}{l}
  {\bar{K}} \left( {\bf x}_1, \ldots, {\bf x}_{2q} \right) = 
  \mathop{\sum}\limits_{\!\!\pi \in \perm \left( \mathbb{N}_{2q} \right)} \mathop{\prod}\limits_{i \in \mathbb{N}_q} K \left( {\bf x}_{\pi_i}, {\bf x}_{\pi_{i+q}} \right)
 \end{array}
\]
This construction does not define a free kernel as the feature map implied by 
$\bar{K}$ depends on $q$.  To construct free kernels we use the following 
theorem (which may be viewed as an analogue of \cite{Smo15} for 
$m$-semi-inner-product-kernels):
\begin{th_innerprod_kern}
 Let $h : \mathbb{R} \to \mathbb{R}$ be a Taylor-expandable function $h (\chi) 
 = \sum_{q} \kappa_q \chi^q$ with $\kappa_q \geq 0$.  Then the function $K_m : 
 (\mathbb{R}^n)^m \to \mathbb{R}$ defined by:
 \[
  \begin{array}{l}
   K_m \left( {\bf x}, {\bf x}' ,\ldots, {\bf x}^{\ldots} \right) = h \left( \ll\! {\bf x}, {\bf x}', \ldots, {\bf x}^{\ldots} \!\gg_m \right)
  \end{array}
 \]
 is a free kernel, where:
 \begin{equation}
  \begin{array}{rl}
   {\mbox{\boldmath $\vartheta$}} \left( {\bf x} \right) &\!\!\!= \mathop{\otimes}\limits_{k_1,k_2,\ldots,k_n \geq 0} x_1^{k_1} x_2^{k_2} \ldots x_n^{k_n} \\
  \end{array}
  \label{eq:standard_feat_form}
 \end{equation}
 \begin{equation}
  \begin{array}{rl}
   {\mbox{\boldmath $\tau$}} &\!\!\!= \mathop{\otimes}\limits_{k_1,k_2,\ldots,k_n \geq 0} \sqrt{\kappa_{(\sum_i k_i)} \left( \begin{array}{c} \sum_i k_i \\ k_1,k_2,\ldots,k_n \\ \end{array} \right)} \\
  \end{array}
  \label{eq:standard_feat_weight}
 \end{equation}
 \label{th:th_innerprod_kern}
\end{th_innerprod_kern}
\begin{proof}
 See supplementary material.
\end{proof}

\begin{table*}
 \centering
 \begin{tabular}{| l || l | l |}
 \hline
   & Free-kernel & Corresponding $2$-kernel \\ 
 \hline
 \hline
  Linear      & $K_m^{\rm lin}  \!\left( {\bf x}, \ldots, {\bf x}^{\ldots} \right) = \;\ll\! {\bf x}, \ldots, {\bf x}^{\ldots} \!\gg_m$                                    & $K^{\rm lin}  \!\left( {\bf x}, {\bf x}' \right) = \left< {\bf x},{\bf x}' \right>$                                              \\ 
  Polynomial  & $K_m^{\rm poly} \!\left( {\bf x}, \ldots, {\bf x}^{\ldots} \right) = \left( 1 + \ll\! {\bf x}, \ldots, {\bf x}^{\ldots} \!\gg_m \right)^q$                 & $K^{\rm poly} \!\left( {\bf x}, {\bf x}' \right) = \left( 1 + \left< {\bf x},{\bf x}' \right> \right)^q$                         \\ 
  Exponential & $K_m^{\rm exp}  \!\left( {\bf x}, \ldots, {\bf x}^{\ldots} \right) = \exp \left( \frac{1}{\sigma} \ll\! {\bf x}, \ldots, {\bf x}^{\ldots} \!\gg_m \right)$ & $K^{\rm exp}  \!\left( {\bf x}, {\bf x}' \right) = \exp \left( \frac{1}{\sigma} \left< {\bf x},{\bf x}' \right> \right)$         \\ 
  SE (RBF)    & $K_m^{\rm se}   \!= \!\exp \!\left( \frac{1}{2\sigma} \left( \!2\! \ll\! {\bf x}, \ldots, {\bf x}^{\ldots} \!\gg_m \!- \left\| {\bf x} \right\|_2 \!\ldots - \left\| {\bf x}^{\ldots} \right\|_2 \right) \right)\!$ & $K^{\rm se}  \!\left( {\bf x}, {\bf x}' \right) = \exp \left( -\frac{1}{2\sigma} \left\| {\bf x}-{\bf x}' \right\|_2^2 \right)$ \\ 
 \hline
 \end{tabular}
 \caption{``Standard'' free and $m$ kernels and their corresponding Mercer 
 ($2$-)kernels.  The linear, polynomial and exponential free kernels are 
 $m$-semi-inner-product free kernels (see theorem \ref{th:th_innerprod_kern}). 
 The SE $m$-kernel $K_m^{\rm se}$ is the normalised form of the 
 exponential free kernel.}
 \label{table:kern_table}
\end{table*}

A sample of free kernels corresponding to $2$-kernels are given in table 
\ref{table:kern_table}.  With the exception of the SE kernel all of 
these kernels are $m$-semi-inner-product kernels; the SE kernel is the 
normalised exponential kernel.

\subsection{Kernel Re-Weighting} \label{sec:kernreweight}

As noted, all free kernels in table \ref{table:kern_table} (except the SE free kernel) have implied 
feature weights defined by (\ref{eq:standard_feat_weight}) and corresponding 
implied features (\ref{eq:standard_feat_form}).  While this is not entirely arbitrary there is no guarantee that the 
weights ${\mbox{\boldmath $\tau$}}$ will give (relatively) more weight to 
important features and (relatively) less to unimportant ones.  We would like 
to be able to {\em adjust} these weights (tune the free kernel) to suit the 
problem at hand.  We call this operation kernel re-weighting.  It relies on 
the following key result:
\begin{th_kernel_weight}
 Let $K_m : (\mathbb{R}^n)^m \to \mathbb{R}$ be a free kernel with implied 
 feature map ${\mbox{\boldmath $\vartheta$}}$ and feature weights 
 ${\mbox{\boldmath $\tau$}}$; and let $\mathcal{E} = \{ ({\bf x}_i,\alpha_i) : 
 x_i \in \mathbb{R}^n, \alpha_i \in \mathbb{R} \}$.  Then the function 
 $K_m^{\mathcal{E}} : (\mathbb{R}^n)^m \to \mathbb{R}$ defined by:
 \begin{equation}
  \!\!\!\!\!\!\begin{array}{l}
   K_m^{\mathcal{E}} \!\left( {\bf x}, \!{\bf x}', \ldots \right) = \sum_{i,j} \!\alpha_i \alpha_j K_{m+2} \!\left( {\bf x}_i, \!{\bf x}_j, \!{\bf x}, \!{\bf x}', \ldots \right)
  \end{array}\!\!\!\!\!\!
  \label{eq:kern_rewight_eq}
 \end{equation}
 is a free kernel family with implied feature map ${\mbox{\boldmath 
 $\vartheta$}}^{\mathcal{E}} = {\mbox{\boldmath $\vartheta$}}$ and implied 
 feature weights ${\mbox{\boldmath $\tau$}}^{\mathcal{E}} = {\bf w} = \sum_i 
 \alpha_i {\mbox{\boldmath $\tau$}}{\mbox{\boldmath $\vartheta$}} ({\bf x}_i)$.
 \label{th:th_kernel_weight}
\end{th_kernel_weight}
\begin{proof}
 Using the definitions:
 \[
  \begin{array}{l}
   K_m^{\mathcal{E}} \left( {\bf x}, \ldots \right) 
    = \sum_{i,j} \alpha_i \alpha_j K_{m+2} \left( {\bf x}_i, {\bf x}_j, {\bf x}, \ldots \right) \\
    = \sum_{i,j} \alpha_i \alpha_j \;\ll\! {\mbox{\boldmath $\tau$}}^2, {\mbox{\boldmath $\vartheta$}} \left( {\bf x}_i \right), {\mbox{\boldmath $\vartheta$}} \left( {\bf x}_j \right), {\mbox{\boldmath $\vartheta$}} \left( {\bf x} \right), \ldots \!\gg_{m+3} \\
    = \;\ll\! \sum_i \alpha_i {\mbox{\boldmath $\tau$}} {\mbox{\boldmath $\vartheta$}} \left( {\bf x}_i \right), \sum_j \alpha_j {\mbox{\boldmath $\tau$}} {\mbox{\boldmath $\vartheta$}} \left( {\bf x}_j \right), {\mbox{\boldmath $\vartheta$}}  \left( {\bf x} \right), \ldots \!\gg_{m+2} \\
    = \;\ll\! {\bf w}^2, {\mbox{\boldmath $\vartheta$}}  \left( {\bf x} \right), \ldots \!\gg_{m+1} \\
  \end{array}
 \]
 and the result follows.
\end{proof}

\begin{figure*}
 \centering 
 \includegraphics[width=\textwidth]{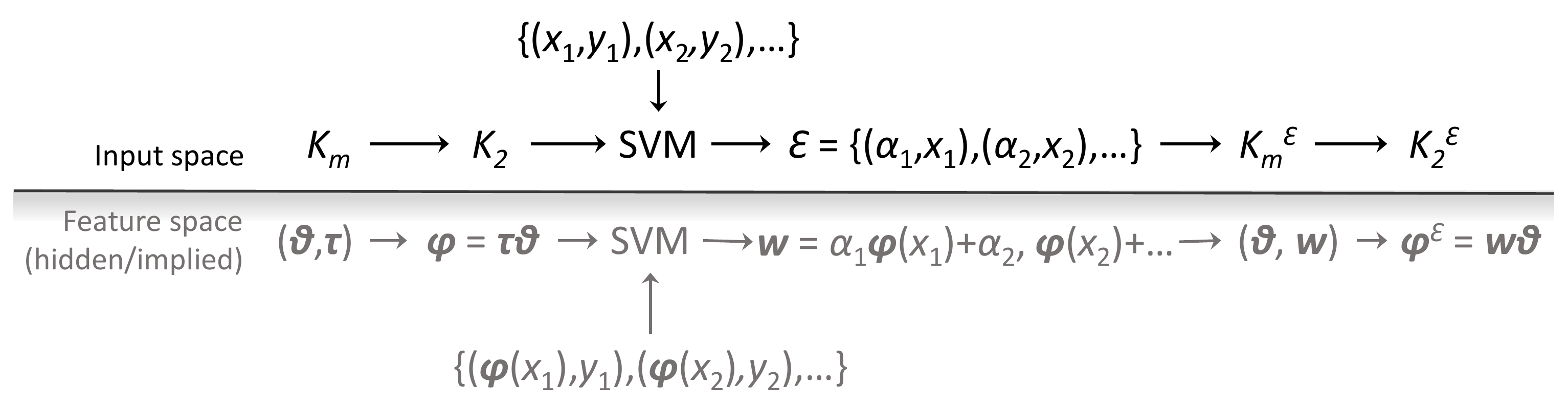} 
 \caption{Kernel Pre-training Procedure.  Working left to right in input 
 space: (1) start with a kernel $K_2$ derived from a free kernel $K_m$.  (2) 
 Train SVM to obtain $\alpha_1, \alpha_2, \ldots$.  (3) Using 
 (\ref{eq:kern_rewight_eq}) obtain re-weighted kernel $K_2^{\mathcal{E}}$ 
 derived from free kernel $K_m^{\mathcal{E}}$.  The feature space operation 
 implied by each step is shown below the line: (1) start with a feature map 
 ${\mbox{\boldmath $\tau$}} {\mbox{\boldmath $\vartheta$}} ({\bf x})$ 
 implied by $K_2$.  (2) Train SVM to obtain weights ${\bf w}$.  (3) Using 
 (\ref{eq:kern_rewight_eq}) obtain re-weighted feature map ${\bf w} 
 {\mbox{\boldmath $\vartheta$}} ({\bf x})$ implied by $K_2^{\mathcal{E}}$.  
 Note that important features (as given by the SVM) $\vartheta_i ({\bf x})$ 
 will be assigned higher weights $w_i$ than less important features.}
 \label{fig:pretrain}
\end{figure*}

\begin{figure}
 \centering 
 \includegraphics[width=7cm]{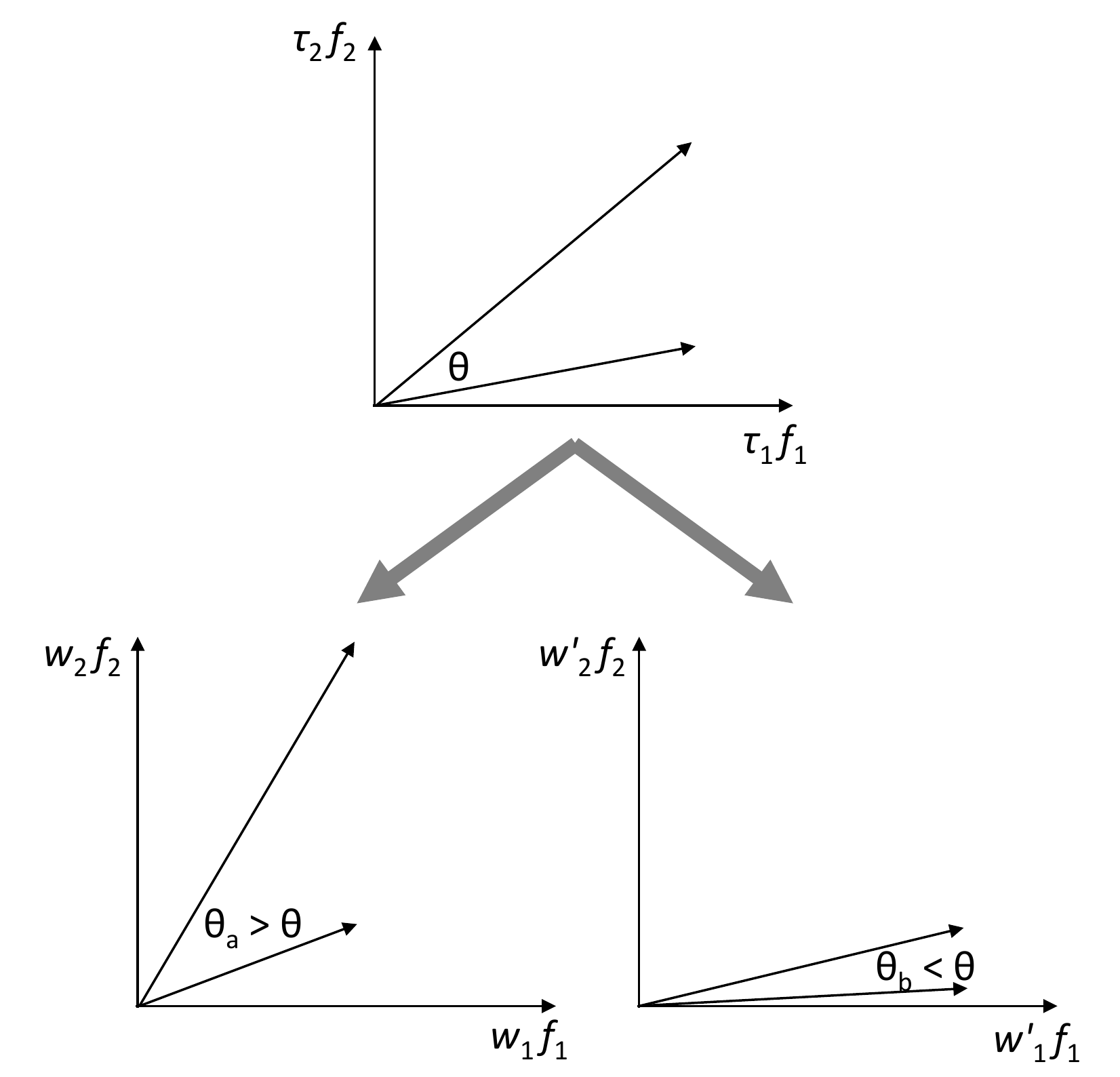} 
 \caption{Geometry of kernel re-weighting ($m = 2$ case).  The upper diagram 
 shows the two vectors ${\mbox{\boldmath $\tau$}} {\mbox{\boldmath 
 $\vartheta$}} ({\bf x})$, ${\mbox{\boldmath $\tau$}} {\mbox{\boldmath 
 $\vartheta$}} ({\bf x}')$ in feature space.  The similarity $K ({\bf x}, {\bf 
 x}')$ as measured by $K$ is proportional to $\cos \theta$ as per 
 (\ref{eq:anglesandstuff}).  The lower diagrams show two possible 
 re-weightings.  The left-hand case increases the relative weight of 
 feature $2$, resulting in decreased similarity, whilst the right-hand case 
 increases the relative weight of feature $1$, resulting in increased 
 similarity.}
 \label{fig:geom_reweigh}
\end{figure}

The utility of this result, as illustrated in figure \ref{fig:pretrain}, is 
that it allows us to take a kernel $K_2$ derived from free kernel $K_m$, train 
a (standard $2$-norm) SVM (or similar) equipped with this kernel to obtain 
(implicitly defined) weights ${\bf w} = \sum_i \alpha_i {\mbox{\boldmath 
$\varphi$}}_2 ({\bf x}_i) = \sum_i \alpha_i {\mbox{\boldmath $\tau$}} 
{\mbox{\boldmath $\vartheta$}} ({\bf x}_i)$ (using (\ref{eq:w_rep}) and 
(\ref{eq:mkernfreemap})), and then use this result to generate a new kernel 
$K_2^{\mathcal{E}}$ derived from the free kernel $K_m^{\mathcal{E}}$ using 
(\ref{eq:kern_rewight_eq})) that gives more weight to important features in 
feature space (as indicated by large $| w_i |$) and less weight to less 
important features (as indicated by small $| w_i |$).  The geometric 
significance of this operation is illustrated in figure 
\ref{fig:geom_reweigh}.

\begin{figure}
 \centering 
 \includegraphics[width=7cm]{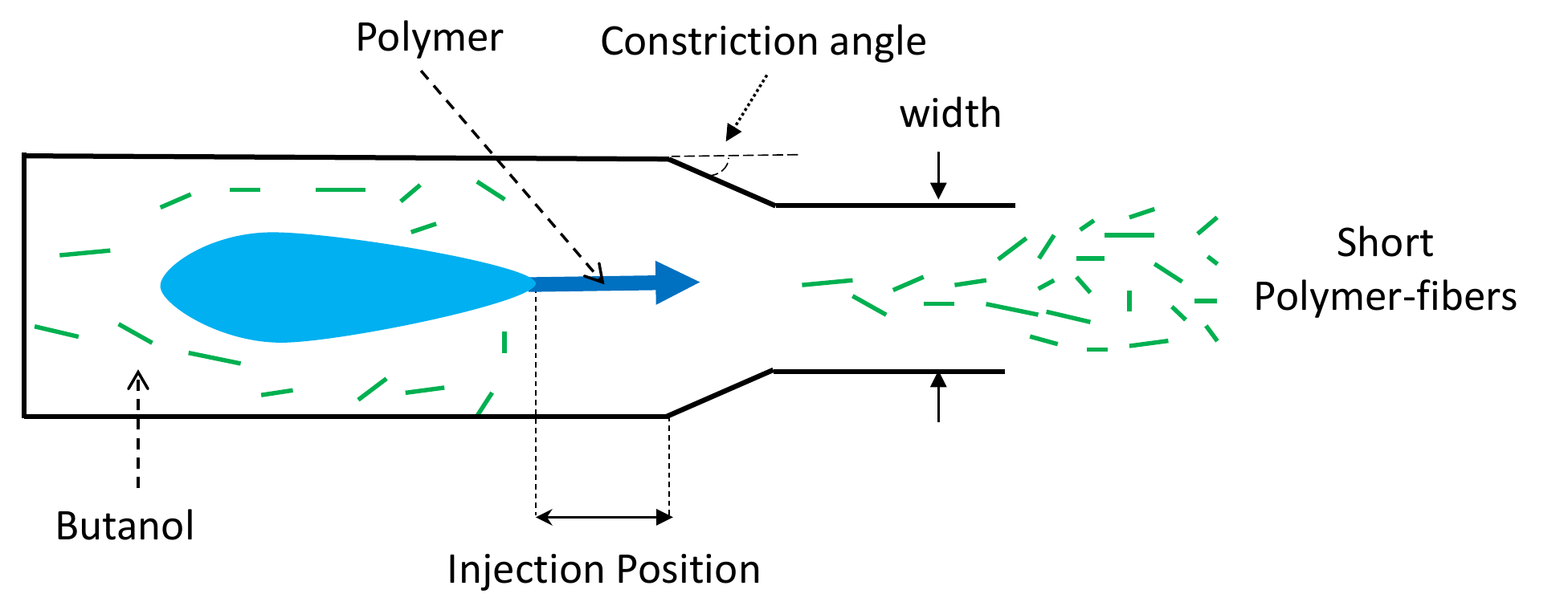} 
 \caption{Device geometry for short polymer fibre injection.}
 \label{fig:devgeom}
\end{figure}

Note that {\em at no point in this process do we need to explicitly know the 
implied feature map or weights}.  Thus we can directly {\em tune} a kernel 
(covariance) function to fit a given set of problem.  As we will show in our 
results, applying this procedure to Bayesian optimisation where the kernel is 
tuned to fit some corpus of condensed knowledge results in a significant 
speed-up due to better fitting of the covariance function.

\section{Proposed Algorithm} \label{sec:modbo}

We now discuss our method for selecting the kernel function.  As noted 
previously, we wish to accelerate Bayesian optimisation by leveraging the 
additional dataset $\mathcal{D}_{\predataset} = \{ ( {\bf x}_{{\predataset} i}, 
y_{{\predataset} i} ) | {\bf x}_{{\predataset} i} \in \mathbb{R}^n, y_{{\predataset} 
i} = f_{\predataset} ( {\bf x}_{{\predataset} i} ) + \epsilon_{\predataset} \in 
\mathbb{Y} \}$, which may be derived for example from relevant patents, 
handbooks or datasheets.  Our proposed algorithm is given in algorithm 
\ref{alg:modded_bbo}.

\begin{algorithm}
\caption{Bayesian Optimisation with Kernel re-weighting.}
\label{alg:modded_bbo}
\begin{algorithmic}
 \INPUT $\mathcal{D}_0, \mathcal{D}_{\predataset}$, free kernel $K_m$.
 \STATE Train SVM with $\mathcal{D}_{\predataset}$ with kernel $K_2$ to obtain 
        $\alpha_{{\predataset}i}$.
 \STATE Construct re-weighted covariance function:
        \[
         \begin{array}{l}
          K_2^{\mathcal{E}} \left( {\bf x}, {\bf x}' \right) 
          = \sum_{i,j} \alpha_{{\predataset}i} \alpha_{{\predataset}j} K_4 \left( {\bf x}_{{\predataset}i}, {\bf x}_{{\predataset}j}, {\bf x}, {\bf x}' \right) \\
         \end{array}
        \]
 \STATE Proceed modelling $f \sim \gp ( 0, K_2^{\mathcal{E}} )$.
 \FOR{$t=1,2,\ldots,T$}
 \STATE Select test point ${\bf x}_t = \argmax_{\bf x} a_t ({\bf x})$.
 \STATE Perform Experiment $y_t = f ( {\bf x}_t ) + \epsilon$.
 \STATE Update $\mathcal{D}_t := \mathcal{D}_{t-1} \cup \{ ({\bf x}_t,y_t) \}$.
 \ENDFOR
\end{algorithmic}
\end{algorithm} 

The key difference between our proposed algorithm and the standard Bayesian 
optimisation algorithm (algorithm \ref{alg:generic_bbo}) is that the 
covariance function $K$ is first re-weighted based on the additional dataset 
$\mathcal{D}_{\predataset}$ to ensure that the implied feature weights match 
those of this dataset before proceeding with the optimisation.  Thus we can 
expect that it will better match the problem at hand in terms of the relative 
weights it gives to components in feature space.

\section{Results} \label{sec:results}

We consider three experiments here, two based on real-world problems and one 
simulation to highlight the properties of the algorithm.

\subsection{Short Polymer Fibres}

In this experiment we have tested our algorithm on the real-world application 
of optimizing short polymer fiber (SPF) to achieve a given (median) target 
length \cite{Li6}.  This process involves the injection of one polymer into 
another in a special device \cite{Sut8}.  The process is controlled by $3$ 
geometric parameters (channel width (mm), constriction angle (degree), device 
position (mm)) and $2$ flow factors (butanol speed (ml/hr), polymer 
concentration (cm/s)) that parametrise the experiment - see figure 
\ref{fig:devgeom}.  Two devices (A and B) were used.  Device A is armed by a 
gear pump and allows for three butanol speeds (86.42, 67.90 and 43.21 cm/s).  The 
newer device B has a lobe pump and allows butonal speed 98, 63 and 48 cm/s.  {\em Our 
goal is to design a new short polymer fiber for Device B that results in a 
(median) target fibre length of 500$\mu$m.}

We write the device parameters as ${\bf x} \in \mathbb{R}^5$ and the result of 
experiments (median fibre length) on each device as $d_{\rm A} ({\bf x})$ and 
$d_{\rm B} ({\bf x})$, respectively.  Device A has been characterised to give 
a dataset $\mathcal{D}_{\predataset} = \{ ({\bf x}_{{\predataset}i}, 
y_{{\predataset}i}) | y_{{\predataset}i} =d_{\rm A} ({\bf x}_{{\predataset}i}) 
\}$ of $163$ input/output pairs.  We aim to minimise:
\[
 \begin{array}{l}
  f \left( {\bf x} \right) = \left( d_{\rm B} \left( {\bf x} \right) - 500 \right)^2
 \end{array}
\]
noting that $f \ne d_{\rm B}$ (the objective $f$ differs from the function 
generating $\mathcal{D}_{\predataset}$, although both relate to fibre length).  
Device B has been similarly characterised and this grid forms our search space 
for Bayesian optimisation.

For this experiment we have used a free SE kernel $K_m^{\rm se}$.  An SVM was 
trained using $\mathcal{D}_{\predataset}$ and an SE kernel (hyperparameters 
$C$ and $\sigma$ were selected to minimise leave-one-out mean-squared-error 
(LOO-MSE)) to obtain $\mathcal{E} = \{ ({\bf x}_{{\predataset}i}, 
\alpha_{{\predataset}i}) \}$.  The re-weighted SE kernel $K_2^{{\rm se} 
\mathcal{E}}$ obtained from this was normalised (to ensure good conditioning 
along the diagonal of ${\bf K}_t$) and used in Bayesian optimisation as per 
algorithm \ref{alg:modded_bbo}.  All data was normalised to $[0,1]$ and all 
experiments were averaged over $40$ repetitions.

\begin{figure}
 \centering 
 \includegraphics[width=7cm]{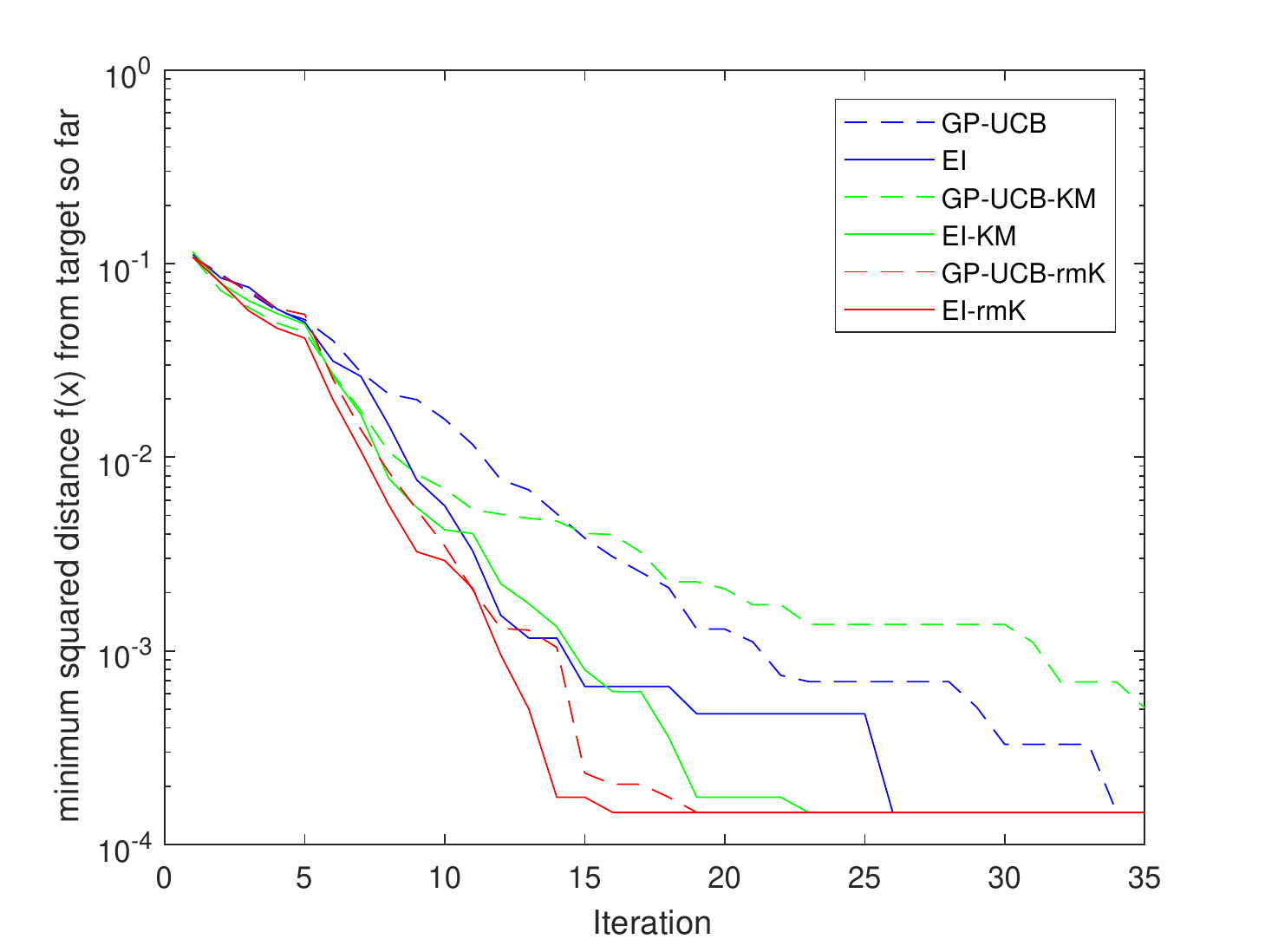} 
 \caption{Short Polymer Fibre design.  Comparison of algorithms in terms of minimum squared distance from the set target versus iterations. GP-UCB and EI indicate acquisition function used.  MK indicates mixture kernel used.  rmK indicates our proposed method.}
 \label{fig:polyres}
\end{figure}

We have tested both the EI and GP-UCB acquisition functions.  Figure 
\ref{fig:polyres} shows the convergence of our proposed algorithm.  Also shown 
for comparison are standard Bayesian optimisation (using a standard SE 
kernel as our covariance function); and a variant of our algorithm where a 
kernel mixture model trained on $\mathcal{D}_{\predataset}$ is used as 
our covariance function - specifically:
\begin{equation}
 \begin{array}{l}
  K \!\left( {\bf x}, {\bf x}' \right) = v_1 K_1 \!\left( {\bf x}, {\bf x}' \right) + v_2 K_2 \!\left( {\bf x}, {\bf x}' \right) + v_3 K_3 \!\left( {\bf x}, {\bf x}' \right)
 \end{array}
 \label{eq:kernmix}
\end{equation}
where $K_1$ is an SE kernel, $K_2$ a Mat{\'e}rn 1/2 kernel, $K_3$ a Mat{\'e}rn 
3/2 kernel; and $v_1, v_2, v_3 \geq 0$ and all relevant (kernel) 
hyperparameters are selected to minimise LOO-MSE on 
$\mathcal{D}_{\predataset}$.  Relevant hyperparameters in Bayesian 
optimisation ($\nu$ for our method and kernel mixtures, $\nu$ and $\sigma$ for 
standard Bayesian optimisation) were selected using max-log-likelihood at each 
iteration.  As can be seen our proposed approach outperforms other methods 
with both GP-UCB and EI acquisition functions.

\subsection{Aluminium Alloy Design using Thermo-Calc}

This experiment considers optimising a new hybrid Aluminium alloy for target yield strength.  
Designing an alloy is an expensive process. Casting an alloy and then measuring its properties usually takes long time.  An alloy has certain phase structures that determine its material 
properties.  For example, phases such as C14LAVES and ALSC3 are known to increase  yield strength whilst others such as  AL3ZR\_D023 and ALLI\_B32 reduce the yield strength of the 
alloy. However a precise function relating the phases to yield strength does not exist.  The simulation software Thermo-Calc 
takes a mixture of component elements as input and 
computes the phase composition of the resulting alloy. We consider 11 elements as potential constituents of the alloy and 24 phases. We use Thermo-Calc for this computation. 

\begin{figure}
 \centering 
 \includegraphics[width=8cm]{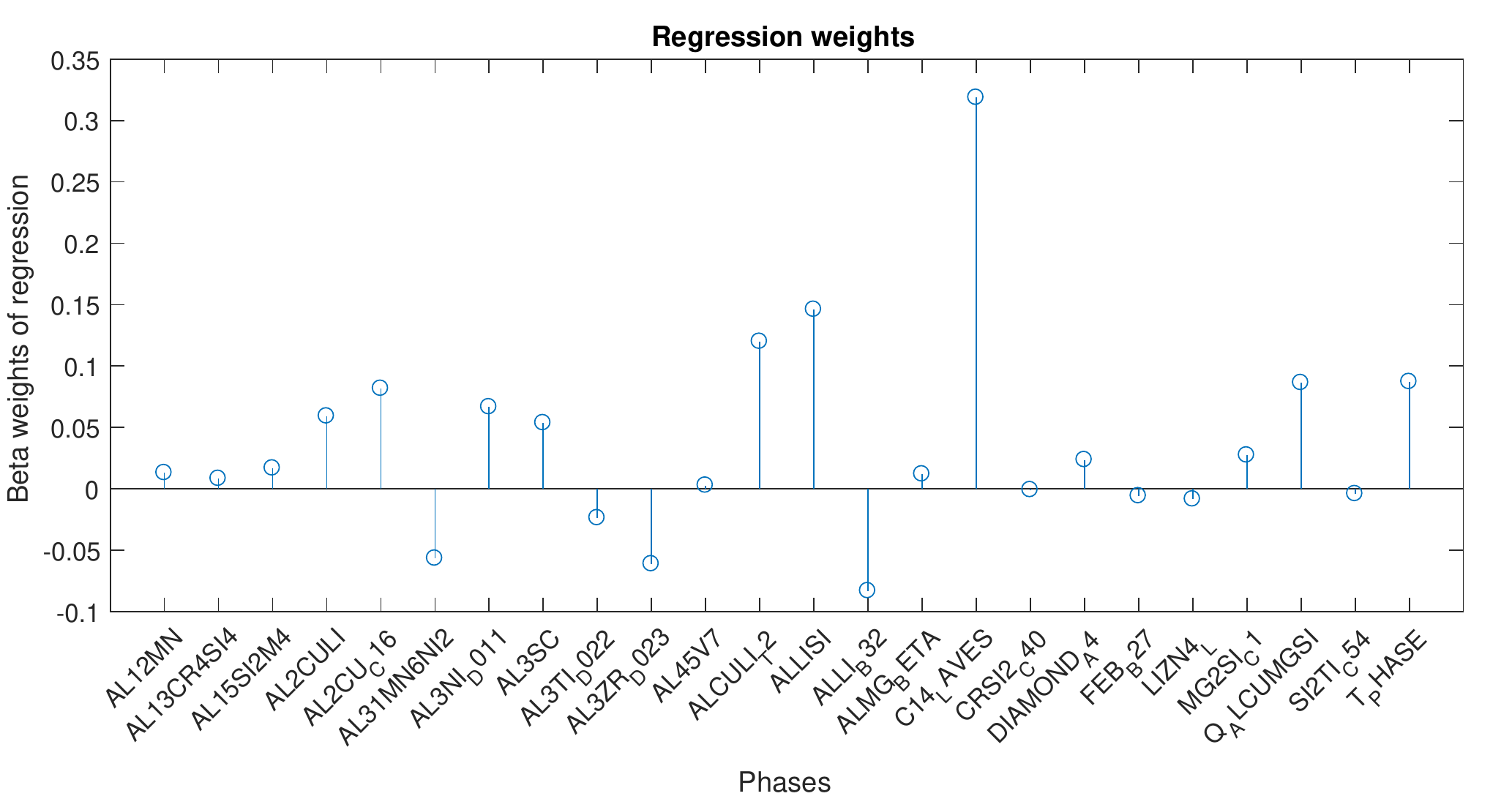} 
 \caption{Regression coefficients for 25 phases as determined from patent data.}
 \label{fig:regcoeff}
\end{figure}

A dataset $\mathcal{D}_{\predataset}$ of $46$ closely related alloys filed as 
patents was collected.  This dataset consists information about the 
composition of the elements in the alloy and their yield strength.  The phase 
compositions extracted from Thermo-Calc simulations for various alloy 
compositions were used to understand the positive or negative contribution of 
phases to the yield strength of the alloy using linear regression.  The 
weights retrieved for these phases were then used formulate a utility 
function.  Figure \ref{fig:regcoeff} shows the regression 
coefficients for the phases contributing to the yield strength.

\begin{figure}
 \centering 
 \includegraphics[width=7cm]{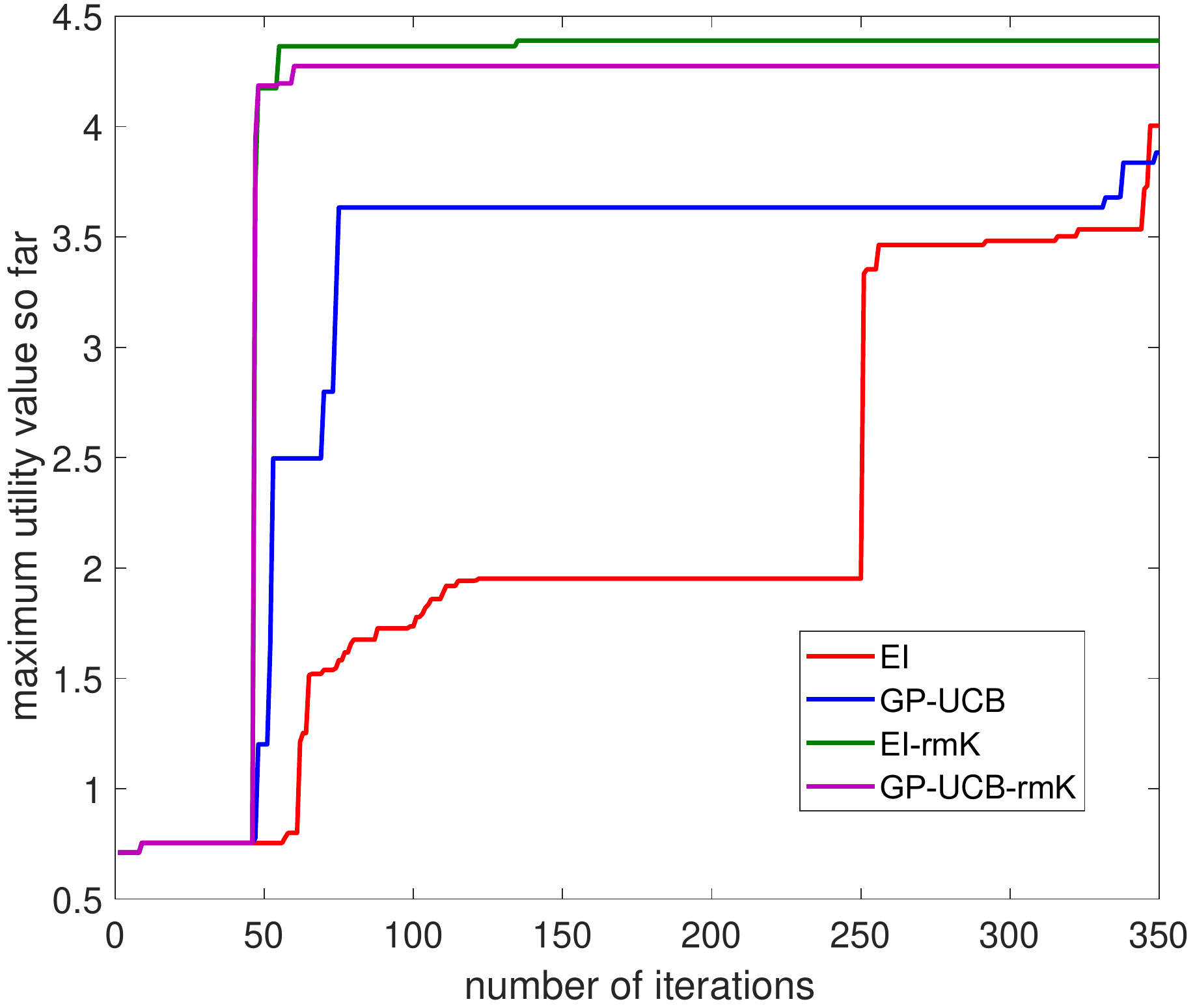} 
 \caption{Aluminium alloy design.  Comparison of algorithms in terms of maximum utility score versus iterations. 
 GP-UCB and EI indicate acquisition function used.  rmK indicates our proposed method.}
 \label{fig:thermores}
\end{figure}

As in the previous experiment for this experiment we have used a free SE 
kernel $K_m^{\rm se}$.  An SVM was trained using $\mathcal{D}_{\predataset}$ 
and an SE kernel (hyperparameters $C$ and $\sigma$ were selected to minimise 
LOO-MSE) to obtain $\mathcal{E} = \{ ({\bf x}_{{\predataset}i}, 
\alpha_{{\predataset}i}) \}$.  The re-weighted SE kernel $K_2^{{\rm se} 
\mathcal{E}}$ obtained from this was normalised (to ensure good conditioning 
along the diagonal of ${\bf K}_t$) and used in Bayesian optimisation as per 
algorithm \ref{alg:modded_bbo}.  All data was normalised to $[0,1]$.

We have tested both the EI and GP-UCB acquisition functions.  Figure 
\ref{fig:thermores} shows the convergence of our proposed algorithm compared 
to standard Bayesian optimisation (using a standard SE kernel as our 
covariance function).  Relevant hyperparameters in Bayesian optimisation 
($\nu$ for our method and kernel mixtures, $\nu$ and $\sigma$ for standard 
Bayesian optimisation) were selected using max-log-likelihood at each 
iteration.  As can be seen our proposed approach outperforms standard Bayesian 
optimisation by a significant margin for both EI and GP-UCB.

\subsection{Simulated Experiment}

\begin{figure}
 \centering 
 \includegraphics[width=7cm]{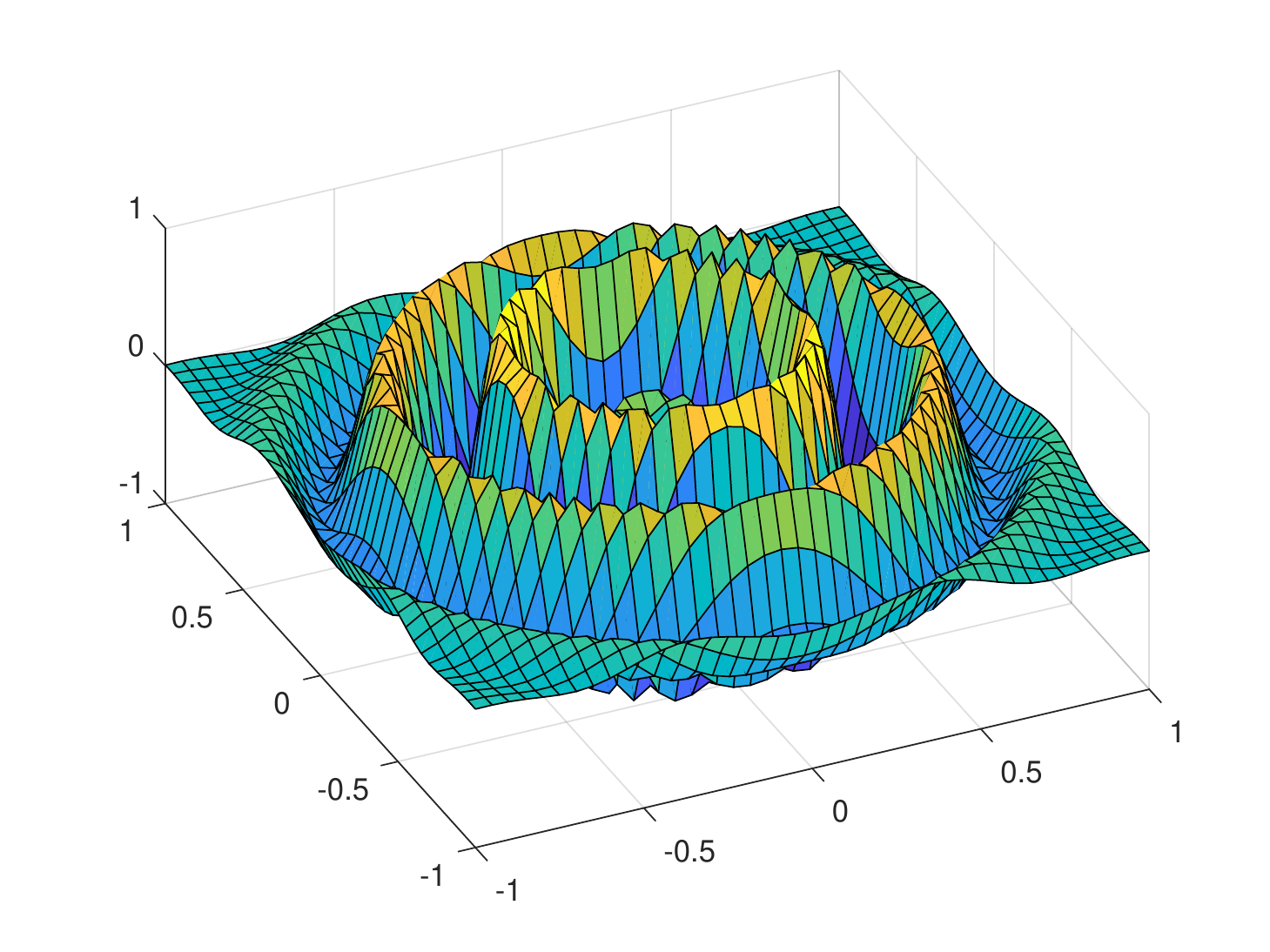} 
 \caption{Target optimisation function in experiment 3.}
 \label{fig:magfunc}
\end{figure}

In this experiment we consider use of kernel re-weighting to incorporate 
domain knowledge into a kernel design.  We aim to minimise the function:
\[
 \begin{array}{l}
  f \left( {\bf x} \right) = 
  \sin \left( 5\pi \left\| {\bf x} \right\|_2 \right) 
  \exp \left( -5 \left( \left\| {\bf x} \right\|_2 - \frac{1}{2} \right)^2 \right)
 \end{array}
\]
as illustrated in figure \ref{fig:magfunc}, where ${\bf x} \in [-1,1]^2$.  
Noting that this function has rotational symmetry we select an additional 
dataset to exhibit this property, namely:
\[
 \begin{array}{l}
  \mathbb{D}_{\predataset} = \left\{ 
  \left. \left( {\bf x}_{{\predataset}i}, y_{{\predataset}i} \right) \right|
  y_{{\predataset}i} = \left\| {\bf x}_{{\predataset}i} \right\|_2
  \right\}
 \end{array}
\]
of $100$ vectors, where ${\bf x}_{{\predataset}i}$ is selected uniformly 
randomly.  Thus $\mathbb{D}_{\predataset}$ reflects the rotational symmetry of 
the target optimisation function $f$ but not its form.

\begin{figure}
 \centering 
 \includegraphics[width=7cm]{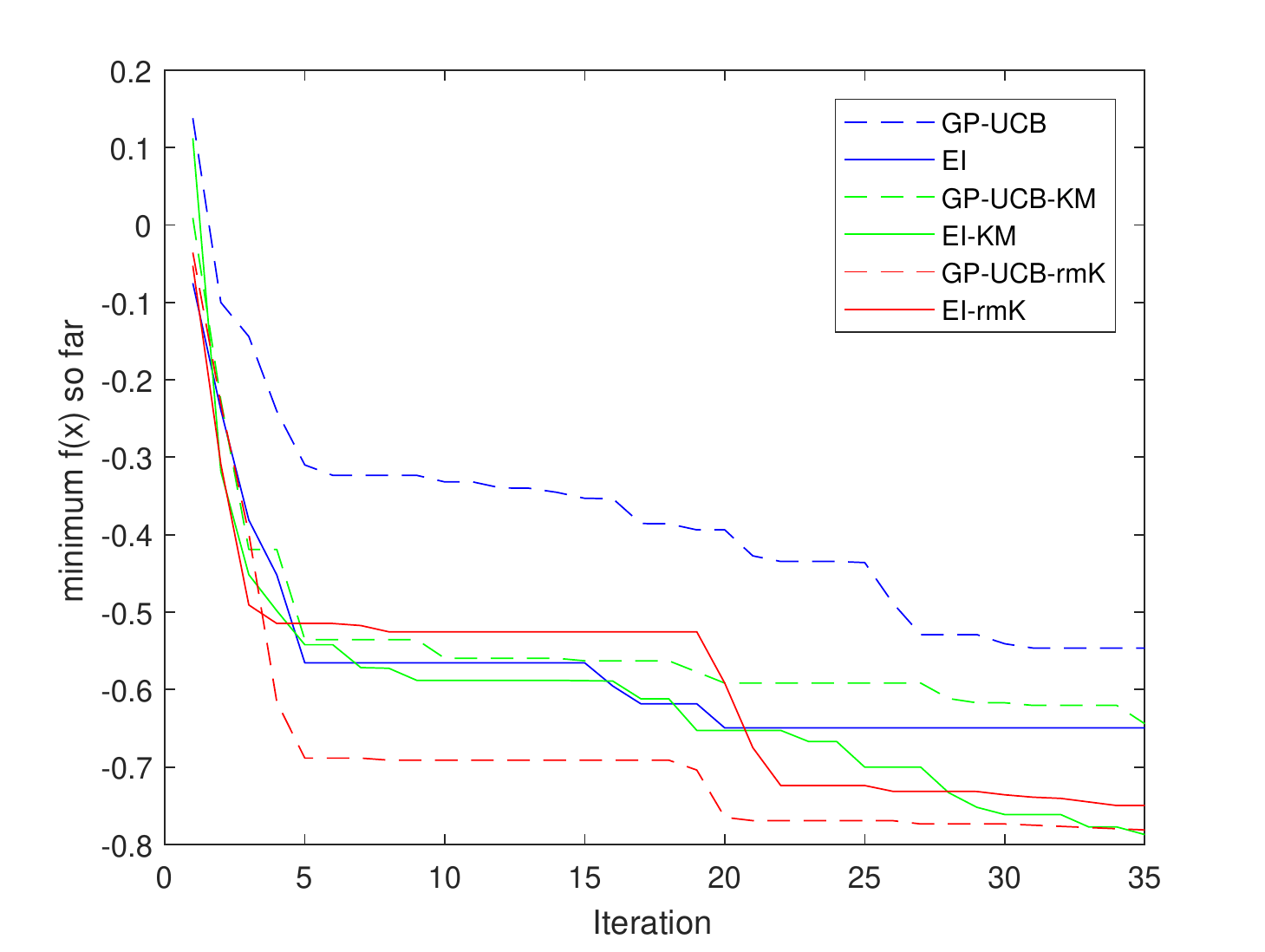} 
 \caption{Simulated experiment.  Comparison of algorithms in terms of minimum $f ({\bf x})$ versus iterations. 
 GP-UCB and EI indicate acquisition function used.  MK indicates mixture kernel used.  rmK indicates our proposed method.}
 \label{fig:mageres}
\end{figure}

As for previous experiments we have started with a free SE kernel $K_m^{\rm 
se}$ and re-weighted this by training and SVM on $\mathcal{D}_{\predataset}$ 
(using an SE kernel with hyperparameters selected to minimise LOO-MSE) to 
obtain $\mathcal{E} = \{ ({\bf x}_{{\predataset}i}, \alpha_{{\predataset}i}) 
\}$, and constructed the re-weighted (but not normalised in this experiment) 
SE kernel $K_2^{{\rm se} \mathcal{E}}$.  However in this case in our 
Bayesian optimisation we have used the composite kernel:
\[
 \begin{array}{l}
  K \left( {\bf x}, {\bf x}' \right) = K_2^{\rm se} \left( K_2^{{\rm se} \mathcal{E}} \left( {\bf x}, {\bf x}' \right) \right)
 \end{array}
\]
which implies a {\em $2$-layer} feature map, the first layer being the 
re-weighted feature map implied by $K_2^{{\rm se} \mathcal{E}}$ and the 
second layer being the standard feature map implied by the SE kernel.  All 
data was normalised to $[0,1]$ and all experiments were averaged over $10$ 
repetitions.

We have tested both the EI and GP-UCB acquisition functions.  Figure 
\ref{fig:mageres} shows the convergence of our proposed algorithm compared to 
standard Bayesian optimisation (using a standard SE kernel) and standard 
Bayesian optimisation with a kernel mixture model as per our short polymer fibre
experiment (\ref{eq:kernmix}) trained on $\mathcal{D}_{\predataset}$ used as 
the covariance function.  Curiously in this case, while our method combined 
with GP-UCB outperforms the alternatives, the results are less clear for our 
method combined with EI.  The precise reason for this will be investigated 
in future work.

\section{Conclusion}

In this paper we have presented a novel approach to direct covariance function 
learning for Bayesian optimisation, focussing on experimental design problems 
where an existing corpus of condensed knowledge is present.  We have based our 
method on $m$-kernel techniques from reproducing kernel Banach space theory.  
We have extended Mercer's theorem to $2q$-kernels; defined free kernel 
families whose implied feature map is independent of $m$ (and provided a means 
of constructing them, with examples); shown how the properties of these may be 
utilised to tune them by effectively modifying the weights in feature space; 
presented a modified Bayesian optimisation algorithm that makes use of these 
properties to pre-tune its covariance function on additional data provided; 
and demonstrated our techniques on two practical optimisation tasks and one 
simulated optimisation problem, demonstrating the efficacy of our proposed 
algorithm.

\section{Supplementary: Derivation of the $p$-norm SVM Dual - Regression Case}

Let $0 < p \leq 2$ be dual to $2q$, $q \in \mathbb{N}^+$ (i.e. $1/p + 1/2q = 
1$).  As defined in the paper, let $\Theta = \{ (x_i,y_i) | x_i \in 
\mathbb{X}, y_i \in \mathbb{R} \}$ be a regression training set 
and let the discriminant function take the standard form:
\[
 \begin{array}{l}
  g \left( x \right) = {\bf w}^\tsp {\mbox{\boldmath $\varphi$}} \left( x \right) + b
 \end{array}
\]
where ${\mbox{\boldmath $\varphi$}}: \mathbb{X} \to \mathbb{R}^d$ is a feature 
map implied by the $2q$-kernel $K : \mathbb{X}^{2q} \to \mathbb{R}$.  
The $p$-norm SVM primal is:
\begin{equation}
 \begin{array}{l}
  \begin{array}{l}
   \mathop{\min}\limits_{{\bf w},b,\xi} \frac{1}{p} \left\| {\bf w} \right\|_p^p + \frac{C}{N} \sum_i \xi_i
  \end{array} \\
  \begin{array}{ll}
   \mbox{such that:} & {\bf w}^\tsp {\mbox{\boldmath $\varphi$}} \left( x_i \right) + b - y_i \leq \epsilon + \xi_i \; \forall i \\
   & {\bf w}^\tsp {\mbox{\boldmath $\varphi$}} \left( x_i \right) + b - y_i \geq -\epsilon - \xi_i \; \forall i \\
   & \xi_i \geq 0 \; \forall i \\
  \end{array} \\
 \end{array}
 \label{eq:theprimal_pnorm_x}
\end{equation}
where $C, \epsilon > 0$.

Defining Lagrange multipliers ${\bar{\beta}}_i \leq 0$, $\beta_i, \gamma_i 
\geq 0$ for, respectively, the first, second and third constraint sets in the 
primal (\ref{eq:theprimal_pnorm_x}) (noting that one or both of ${\bar{\beta}}_i$ and 
$\beta_i$ must be zero for all $i$, as at most one the first two constraints 
can be met with equality for any $i$) we obtain the Lagrangian:
\begin{equation}
 \!\!\!\!\!\!\begin{array}{rl}
  L &\!\!\!= \frac{1}{p} \left\| {\bf w} \right\|_p^p + \frac{C}{N} \sum_i \xi_i - \sum_i {\bar{\beta}}_i {\bf w}^\tsp {\mbox{\boldmath $\varphi$}} \left( x_i \right) - \ldots \\
  & \ldots - \sum_i {\bar{\beta}}_i b + \sum_i {\bar{\beta}}_i y_i + \epsilon \sum_i {\bar{\beta}}_i + \ldots \\
  & \ldots + \sum_i {\bar{\beta}}_i \xi_i - \sum_i \beta_i {\bf w}^\tsp {\mbox{\boldmath $\varphi$}} \left( x_i \right) - \ldots \\
  & \ldots - \sum_i \beta_i b + \sum_i \beta_i y_i - \epsilon \sum_i \beta_i - \ldots \\
  & \ldots - \sum_i \beta_i \xi_i - \sum_i \gamma_i \xi_i \\
 \end{array}\!\!\!\!\!\!
 \label{eq:thelagrangian}
\end{equation}
Using the fact that one or both of $\beta_i \geq 0$ and ${\bar{\beta}}_i 
\leq 0$ must be zero for all $i$ we see that ${\bar{\beta}}_i - \beta_i = - | 
{\bar{\beta}}_i + \beta_i |$ for all $i$, and hence:
\[
 \begin{array}{l}
  \nabla_{\bf w} L = \sgn \left( {\bf w} \right) \left| {\bf w} \right|^{p-1} - \sum_i \left( {\bar{\beta}}_i + \beta_i \right) {\mbox{\boldmath $\varphi$}} \left( x_i \right) \\
  \frac{\partial L}{\partial b} = - \sum_i \left( {\bar{\beta}}_i + \beta_i \right) \\
  \frac{\partial L}{\partial \xi_i}  = \frac{C}{N} - \left| {\bar{\beta}}_i + \beta_i \right| - \gamma_i \\
 \end{array}
\]
where $|{\bf w}|$ and ${\bf w}^{p-1}$ are, respectively, the elementwise 
absolute value and elementwise power of ${\bf w}$.  It follows that, for 
optimality:
\[
 \begin{array}{l}
  {\bf w} = \sum_i \left( {\bar{\beta}}_i + \beta_i \right) \sgn \left( {\mbox{\boldmath $\varphi$}} \left( x_i \right) \right) \left| {\mbox{\boldmath $\varphi$}} \left( x_i \right) \right|^{2q-1} \\
  \sum_i \left( {\bar{\beta}}_i + \beta_i \right) = 0 \\
  -\frac{C}{N} \leq \left( {\bar{\beta}}_i + \beta_i \right) \leq \frac{C}{N} \; \forall i \\
 \end{array}
\]

Defining $\alpha_i = {\bar{\beta}}_i + \beta_i$ $\forall i$, substituting into 
the Lagrangian and using the definition of the $2q$-kernel $K$ (the 
$2q$-kernel trick) we obtain the dual $p$-norm SVM training problem:
\begin{equation}
 \!\!\!\!\!\!\!\!\!\!\!\!\begin{array}{l}
  \begin{array}{rl}
   \mathop{\min}\limits_{\alpha} &\!\!\!\frac{1}{2q} \mathop{\sum}\limits_{i_1,i_2,\ldots,i_{2q}} \!\!\!\alpha_{i_1} \alpha_{i_2} \ldots \alpha_{i_{2q}} K_{i_1, i_2, \ldots, i_{2q}} + \ldots \\
                                 &\!\!\!\ldots + \epsilon \sum_i \left| \alpha_i \right| - \sum_i y_{i} \alpha_i
  \end{array} \\
  \begin{array}{ll}
   \mbox{such that:} & -\frac{C}{N} \leq \alpha_i \leq \frac{C}{N} \; \forall i \\
   & \sum_i \alpha_i = 0 \\
  \end{array} \\
 \end{array}\!\!\!\!\!\!\!\!\!\!\!\!
 \label{eq:thedual_x}
\end{equation}
where $K_{i_1, i_2, \ldots, i_{2q}} = K ( x_{i_1}, x_{i_2}, \ldots, 
x_{i_{2q}} )$; and the discriminant function may be written:
\begin{equation}
 \begin{array}{l}
  g \left( x \right) = \!\!\!\!\!\mathop\sum\limits_{i_2, \ldots, i_{2q}}\!\!\!\!\! \alpha_{i_2} \ldots \alpha_{i_{2q}} K \left( x, x_{i_2}, \ldots, x_{i_{2q}} \right) + b
 \end{array}
 \label{eq:dual_p_trained_x}
\end{equation}

\section{Supplementary: Derivation of the $p$-norm SVM Dual - Binary Classification Case}

In the case of binary classification $\Theta = \{ (x_i,y_i) | x_i \in 
\mathbb{X}, y_i \in \{ -1,+1 \} \}$.  The discriminant function is:
\[
 \begin{array}{l}
  g \left( x \right) = {\bf w}^\tsp {\mbox{\boldmath $\varphi$}} \left( x \right) + b
 \end{array}
\]
where ${\mbox{\boldmath $\varphi$}}: \mathbb{X} \to \mathbb{R}^d$ is a feature 
map implicit in the $2q$-kernel $K : \mathbb{X}^{2q} \to \mathbb{R}$.  
The $p$-norm SVM primal is:
\begin{equation}
 \begin{array}{l}
  \begin{array}{l}
   \mathop{\min}\limits_{{\bf w},b,\xi} \frac{1}{p} \left\| {\bf w} \right\|_p^p + \frac{C}{N} \sum_i \xi_i
  \end{array} \\
  \begin{array}{ll}
   \mbox{such that:} & y_i \left( {\bf w}^\tsp {\mbox{\boldmath $\varphi$}} \left( x_i \right) + b \right) \geq 1 - \xi_i \; \forall i \\
   & \xi_i \geq 0 \; \forall i \\
  \end{array} \\
 \end{array}
 \label{eq:theprimal_pnorm_cls}
\end{equation}

Defining Lagrange multipliers ${\bar{\alpha}}_i, \gamma_i \geq 0$ for, 
respectively, the first and second constraint sets in the primal 
(\ref{eq:theprimal_pnorm_cls}) we obtain the Lagrangian:
\begin{equation}
 \!\!\!\!\!\!\begin{array}{rl}
  L &\!\!\!= \frac{1}{p} \left\| {\bf w} \right\|_p^p + \frac{C}{N} \sum_i \xi_i - \sum_i {\bar{\alpha}}_i y_i {\bf w}^\tsp {\mbox{\boldmath $\varphi$}} \left( x_i \right) - \ldots \\
  & \ldots - \sum_i {\bar{\alpha}}_i y_i b + \sum_i {\bar{\alpha}}_i - \sum_i {\bar{\alpha}}_i \xi_i - \sum_i \gamma_i \xi_i \\
 \end{array}\!\!\!\!\!\!
 \label{eq:thelagrangian_cls}
\end{equation}
and hence:
\[
 \begin{array}{l}
  \nabla_{\bf w} L = \sgn \left( {\bf w} \right) \left| {\bf w} \right|^{p-1} - \sum_i {\bar{\alpha}}_i y_i {\mbox{\boldmath $\varphi$}} \left( x_i \right) \\
  \frac{\partial L}{\partial b} = \sum_i y_i {\bar{\alpha}}_i \\
  \frac{\partial L}{\partial \xi_i}  = \frac{C}{N} - {\bar{\alpha}}_i - \gamma_i \\
 \end{array}
\]
where $|{\bf w}|$ and ${\bf w}^{p-1}$ are, respectively, the elementwise 
absolute value and elementwise power of ${\bf w}$.  It follows that, for 
optimality:
\[
 \begin{array}{l}
  {\bf w} = \sum_i {\bar{\alpha}}_i y_i \sgn \left( {\mbox{\boldmath $\varphi$}} \left( x_i \right) \right) \left| {\mbox{\boldmath $\varphi$}} \left( x_i \right) \right|^{2q-1} \\
  \sum_i y_i {\bar{\alpha}}_i = 0 \\
  0 \leq {\bar{\alpha}}_i \leq \frac{C}{N} \; \forall i \\
 \end{array}
\]

Defining $\alpha_i = y_i {\bar{\alpha}}_i$ $\forall i$, substituting into the 
Lagrangian and using the definition of the $2q$-kernel $K$ (the 
$2q$-kernel trick) we obtain the dual $p$-norm SVM training problem:
\begin{equation}
 \!\!\!\!\!\!\!\!\!\!\!\!\begin{array}{l}
  \begin{array}{l}
   \mathop{\min}\limits_{\alpha} \frac{1}{2q} \mathop{\sum}\limits_{i_1,i_2,\ldots,i_{2q}} \!\!\!\alpha_{i_1} \alpha_{i_2} \ldots \alpha_{i_{2q}} K_{i_1, i_2, \ldots, i_{2q}} - \sum_i y_i \alpha_i
  \end{array} \\
  \begin{array}{ll}
   \mbox{such that:} & 0 \leq y_i \alpha_i \leq \frac{C}{N} \; \forall i \\
   & \sum_i \alpha_i = 0 \\
  \end{array} \\
 \end{array}\!\!\!\!\!\!\!\!\!\!\!\!
 \label{eq:thedual_cls}
\end{equation}
where $K_{i_1, i_2, \ldots, i_{2q}} = K ( x_{i_1}, x_{i_2}, \ldots, 
x_{i_{2q}} )$; and the discriminant function may be written:
\begin{equation}
 \begin{array}{l}
  g \left( x \right) = \!\!\!\!\!\mathop\sum\limits_{i_2, \ldots, i_{2q}}\!\!\!\!\! \alpha_{i_2} \ldots \alpha_{i_{2q}} K \left( x, x_{i_2}, \ldots, x_{i_{2q}} \right) + b
 \end{array}
 \label{eq:dual_p_trained_cls}
\end{equation}

\section{Supplementary: Proof of Theorem 2}

Mercer's theorem states that \cite{Mer1}:
\begin{mercer_theorem}[Mercer's theorem]
 Let $K : \mathbb{X} \times \mathbb{X} \to \mathbb{R}$ be a continuous function 
 for which $K (x,x') = K (x',x)$ and:
 \[
  \begin{array}{l}
   \int_{x,x' \in \mathbb{X}} K \left( x,x' \right) \psi \left( x \right) \psi \left( x' \right) dx dx' \geq 0
  \end{array}
 \]
 for all $\psi : \mathbb{X} \to \mathbb{R}$ satisfying $\int_{x \in \mathbb{X}} 
 |\psi(x)|^2 dx < \infty$.  Then there exists ${\mbox{\boldmath $\varphi$}} : 
 \mathbb{X} \to \mathbb{R}^d$, $d \in \mathbb{N}^+ \cup \{ \infty \}$ such that:
 \[
  \begin{array}{l}
   K \left( x,x' \right) = \left< {\mbox{\boldmath $\varphi$}} \left( x \right) {\mbox{\boldmath $\varphi$}} \left( x' \right) \right>
  \end{array}
 \]
 the series being uniformly and absolutely convergent. 
 \label{th:mercer_theorem}
\end{mercer_theorem}

We now prove theorem 3 from the paper:
\begin{proof}
It follows from theorem~\ref{th:mercer_theorem} that for any function $K$ 
satisfying the conditions given in the theorem the function $L$ defined by 
$K$ may be written as:
\[
 \begin{array}{l}
  L \left( z, z' \right) = \sum_n \nu_n \left( z \right) \nu_n \left( z' \right)
 \end{array}
\]
the series being uniformly and absolutely convergent.  Hence:
\[
 \begin{array}{l}
  K \left( x_1, \ldots, x_{2q} \right) = \ldots \\
  \; \ldots = \sum_n \kappa_n \left( x_1, \ldots, x_q \right) \kappa_n \left( x_{q+1}, \ldots, x_{2q} \right)
 \end{array}
\]
where $\kappa_n ( x_1, \ldots, x_q ) = \nu_n ( [ x_1, \ldots, x_q ] )$.  If $q 
= 1$ this is sufficient to prove the theorem.  For the more general case we 
first note that, using the symmetry of $K$:
\begin{equation}
 \!\!\!\!\!\!\begin{array}{l}
    \kappa_n \!\left( x_1, \ldots, x_i, \ldots, x_q \right) \kappa_n \!\left( x_{q+1}, \ldots, x_j, \ldots, x_{2q} \right) \\ 
  = \kappa_n \!\left( x_1, \ldots, x_j, \ldots, x_q \right) \kappa_n \!\left( x_{q+1}, \ldots, x_i, \ldots, x_{2q} \right)
 \end{array}\!\!\!\!\!\!
\label{eq:switch_args_a}
\end{equation}
for all $i \leq q < j \in \mathbb{N}_{2q}$ (we denote $\mathbb{N}_r = \{ 1,2,
\ldots,r \}$ ).  This means that for any product of two such $\kappa_n$ 
functions we may switch arguments between the two $\kappa_n$ functions without 
changing the product.  Changing the range of $j$ in the above reasoning to $i 
< j \leq q$ we see that for all $i < j \in \mathbb{N}_q$:
\begin{equation}
 \begin{array}{l}
    \kappa_n \!\left( x_1, \ldots, x_i, \ldots, x_j \ldots x_q \right) \\
  = \kappa_n \!\left( x_1, \ldots, x_j, \ldots, x_i \ldots, x_q \right)
 \end{array}
\label{eq:switch_args_b}
\end{equation}

Next suppose that $\exists n$ such that $\kappa_n ( a,\ldots,a) = 0$ $\forall 
a \in \mathbb{X}$.  Using (\ref{eq:switch_args_a}) and 
(\ref{eq:switch_args_b}) we see that $\forall a_1,\ldots,a_q \in \mathbb{X}$:
\[
 \begin{array}{l}
  \kappa_n^q \left( a_1, \ldots, a_q \right)
  = \prod_i \kappa_n \left( a_i, \ldots, a_i \right) = 0 \\
 \end{array}
\]
which implies that $\kappa_n$ is trivially zero everywhere and may therefore 
be removed from the series.  Hence we may assume that $\forall n$ there exists 
$a_n \in \mathbb{X}$ such that $\kappa_n (a_n,\ldots,a_n) \ne 0$.  It further 
follows from (\ref{eq:switch_args_a}), (\ref{eq:switch_args_b}) that $\forall 
r < q$:
\[
 \begin{array}{r}
 \kappa_n ( x_1, \ldots, x_r, x_{r+1}, \underbrace{a_n, \ldots, a_n}_{q-r-1} ) \kappa_n \left( a_n, \ldots, a_n \right) = \ldots \\
 \ldots = \kappa_n ( x_1, \ldots, x_r, \underbrace{a_n, \ldots, a_n}_{q-r} ) \kappa_n \left( x_{r+1}, a_n, \ldots, a_n \right)
 \end{array}
\]
and therefore:
\begin{equation}
 \begin{array}{r}
  \kappa_n ( x_1, \ldots, x_r, x_{r+1}, \underbrace{a_n, \ldots, a_n}_{q-r-1} ) = \ldots \\
  \ldots = \kappa_n ( x_1, \ldots, x_r, \underbrace{a_n, \ldots, a_n}_{q-r} ) \phi_n \left( x_{r+1} \right)
 \end{array}
\label{eq:mercer_keyb}
\end{equation}
where:
\[
\begin{array}{l}
 \phi_n \left( x \right) = \frac{\kappa_n \left( x, a_n, \ldots, a_n \right)}{\kappa_n \left( a_n, \ldots, a_n \right)} \\
\end{array}
\]

Recursively applying (\ref{eq:mercer_keyb}) to $\kappa_n ( x_1, \ldots, x_q )$ 
it may be seen that:
\[
\begin{array}{l}
 \kappa_n \left( x_1, \ldots, x_q \right) = \kappa_n \left( a_n, \ldots, a_n \right) \phi_n \left( x_1 \right) \ldots \phi_n \left( x_q \right)
\end{array}
\]
from which we may deduce that there exists a uniformly and absolutely 
convergent series representation for $K$:
\[
 \begin{array}{c}
  K \left( x_1, \ldots, x_{2q} \right) = \sum_n \varphi_n \left( x_1 \right) \ldots \varphi_n \left( x_{2q} \right)
 \end{array}
\]
where:
\[
 \begin{array}{l}
  \varphi_n \left( x \right) = \frac{\kappa_n \left( x, a_n, \ldots, a_n \right)}{\left| \kappa_n \left( a_n, \ldots, a_n \right) \right|^{\frac{q-1}{q}}}
 \end{array}
\]
and $a_n \in \{ a \in \mathbb{X} \subset \mathbb{R}^d | \kappa_n ( a, \ldots, 
a ) \ne 0 \} \ne \emptyset$ $\forall n$, which completes the proof.
\end{proof}

\section{Supplementary: Proof of Theorem 3}

\begin{proof}
Using the multinomial theorem it may be seen that:
\[
 \begin{array}{l}
  h \left( \ll\! {\bf x}, \ldots, {\bf x}^{\ldots} \!\gg_m \right) 
  = \mathop{\sum}\limits_{q \geq 0} \kappa_q \ll\! {\bf x}, \ldots, {\bf x}^{\ldots} \!\gg_m^q \\
  = \mathop{\sum}\limits_{k_1,k_2,\ldots,k_n \geq 0} \tau^2_{m,k_{\bullet}} \theta_{m,k_{\bullet}} \left( {\bf x} \right) \ldots \theta_{m,k_{\bullet}} \left( {\bf x}^{\ldots} \right) \\
 \end{array}
\]
where:
\[
 \begin{array}{l}
  \theta_{m,k_{\bullet}} \left( {\bf x} \right) = \mathop{\prod}\limits_{i \in \mathbb{N}_n}\! \left( x_i \right)^{k_i} \\
  \tau_{m,k_{\bullet}} = \sqrt{\kappa_{(\sum_i k_i)} \left( \begin{array}{c} \sum_i k_i \\ k_1,k_2,\ldots,k_n \\ \end{array} \right)} \\
 \end{array}
\]
and hence:
\[
 \begin{array}{l}
  K_m \left( {\bf x}, \ldots {\bf x}^{\ldots} \right) 
  = \;\ll\! {\mbox{\boldmath $\tau$}}^2, {\mbox{\boldmath $\vartheta$}} \left( {\bf x} \right), \ldots, {\mbox{\boldmath $\vartheta$}} \left( {\bf x}^{\ldots} \right) \!\gg_{m+1}
 \end{array}
\]
where 
 \[
  \begin{array}{rl}
   {\mbox{\boldmath $\vartheta$}} \left( {\bf x} \right) &\!\!\!= \mathop{\otimes}\limits_{k_1,k_2,\ldots,k_n \geq 0} x_1^{k_1} x_2^{k_2} \ldots x_n^{k_n} \\
  \end{array}
 \]
and:
 \[
  \begin{array}{rl}
   {\mbox{\boldmath $\tau$}} &\!\!\!= \mathop{\otimes}\limits_{k_1,k_2,\ldots,k_n \geq 0} \sqrt{\kappa_{(\sum_i k_i)} \left( \begin{array}{c} \sum_i k_i \\ k_1,k_2,\ldots,k_n \\ \end{array} \right)} \\
  \end{array}
 \]
are independent of $m$.
\end{proof}

\bibliographystyle{plain}
\bibliography{universal}

\end{document}